\documentclass[conference]{IEEEtran}
\pdfoutput=1 % arXiv: force pdfLaTeX (JPG figures + pgfplots need the PDF path)

\usepackage[letterpaper,
            left=0.625in,
            right=0.625in,
            top=0.75in,
            bottom=1in]{geometry}

\usepackage{graphicx}
\usepackage{amsmath}
\usepackage{amssymb}
\usepackage{booktabs}
\usepackage{array}
\usepackage{multirow}
\usepackage{subcaption}
\usepackage{pgfplots}
\usepackage{xcolor}
\usepackage{algorithm}
\usepackage{algpseudocode}
\usepackage{cite}
\usepackage{hyperref}
\usepackage{xurl}
\pgfplotsset{compat=1.17}

\hypersetup{
    colorlinks=true,
    urlcolor=blue,
    linkcolor=black,
    citecolor=black,
    filecolor=black,
    pdftitle={Visual Forensic Detection of Fabricated Bangla News Photocards using Deep Learning},
    pdfauthor={Anindya Kundu, Kabya Mithun Saha, Md. Muhaiminul Islam, Fayek Ahmed, Aminul Kader Bulbul, Muhammad Ibrahim}
}

\makeatletter
\newcommand{\linebreakand}{%
  \end{@IEEEauthorhalign}
  \hfill\mbox{}\par\vspace{-4pt}
  \mbox{}\hfill\begin{@IEEEauthorhalign}
}
\makeatother

\begin{document}

\title{Visual Forensic Detection of Fabricated Bangla News Photocards using Deep Learning}

\author{
\IEEEauthorblockN{Anindya Kundu}
\IEEEauthorblockA{anindya-2021911185@cs.du.ac.bd}
\and
\IEEEauthorblockN{Kabya Mithun Saha}
\IEEEauthorblockA{kabyamithun-2021111192@cs.du.ac.bd}
\and
\IEEEauthorblockN{Md. Muhaiminul Islam}
\IEEEauthorblockA{muhaiminulislam-2021111219@cs.du.ac.bd}
\linebreakand
\IEEEauthorblockN{Fayek Ahmed}
\IEEEauthorblockA{fayek-2021511233@cs.du.ac.bd}
\and
\IEEEauthorblockN{Aminul Kader Bulbul}
\IEEEauthorblockA{aminulkader@du.ac.bd}
\linebreakand
\IEEEauthorblockN{Muhammad Ibrahim\textsuperscript{*}}
\IEEEauthorblockA{ibrahim313@du.ac.bd}
\linebreakand
\IEEEauthorblockA{\rule{0pt}{18pt}Department of Computer Science and Engineering\\
University of Dhaka, Dhaka-1000, Bangladesh\\[4pt]
\textsuperscript{*}Corresponding author}
}

\maketitle

% ---------------------------------------------------------------
\begin{abstract}
In Bangladesh, a large share of localized misinformation spreads as \textit{photocards}: single-image news summaries that anyone can fake and that copy the look and branding of well-known news organizations. The trouble is that the misleading message lives in the pixels rather than in machine-readable text, so the natural-language fact-checking tools we normally lean on never even see it. In this work we recast fake-news detection for the photocard as a problem of visual image forgery, and we ask how far a small, purpose-built convolutional model can take us.

To study the problem we built a localized dataset of 969 authentic and 1,077 fabricated Bangla photocards, all of it scraped from public Facebook pages: the authentic cards come from the verified pages of seventeen national television broadcasters, while the fabricated ones were taken from the pseudo-broadcaster pages, anonymous profiles, and public groups that manufacture and spread them. Images shared online seldom stay pristine, so to mimic the photometric and morphological wear they accumulate in circulation we applied sixteen different transformations to each base image, growing the corpus to 15,504 authentic and 17,232 fabricated samples (32,736 in all); every file was then passed through a TensorFlow decode pass to strip out byte-level corruption. Our starting point is deliberately weak. A fully connected Artificial Neural Network (ANN) manages only about 52\% accuracy, its loss stuck at the binary-entropy floor $\ln 2 \approx 0.693$. That near-random behaviour is itself the evidence: once an image is flattened, the spatial correlations forgery detection relies on are simply gone. Against this we set a compact, fully batch-normalized four-block Convolutional Neural Network (CNN) with 489,954 parameters running at $256\times256$ pixels---roughly $1{,}644\times$ smaller than the ANN. We are careful about how we measure it, adopting \textbf{leakage-free stratified group five-fold cross-validation}: every augmented version of a given card is tied to one fold, so the model never trains on an augmentation of an image it is later tested on. Under this protocol the CNN reaches \textbf{87.45\%\,$\pm$\,0.53\%} pooled cross-validated accuracy from a random start, climbing to \textbf{88.75\%\,$\pm$\,1.55\%} once its classifier head is warm-started from the ANN. As a strong pretrained reference, an ImageNet-pretrained \textbf{YOLOv8n-cls} backbone ($1.44$M parameters) fine-tuned under the identical protocol reaches \textbf{97.36\%\,$\pm$\,0.42\%}, marking the headroom that still sits above the compact, purpose-built model. For the two CNNs and the YOLOv8 baseline we report per-class precision, recall, $F_1$, and pooled confusion matrices. Along the way we describe a trap we fell into and how we got out of it: leaving the classifier head unnormalized drives the network to collapse into predicting a single class, a failure that batch-normalizing the head---with gradient clipping and a reduced learning rate---reliably cures. The paper's contributions are the dataset, the preprocessing pipeline, the negative ANN result that motivates spatial modelling, two carefully cross-validated CNN benchmarks alongside a pretrained YOLOv8-cls reference, a manual error analysis and bias--variance diagnosis, and a leakage-aware evaluation protocol for OCR-free photocard verification.
\end{abstract}

\begin{IEEEkeywords}
Image forgery detection, Bangla misinformation, photocards, deep learning, convolutional neural networks, transfer learning, pretrained backbones, YOLOv8, group cross-validation, social media forensics.
\end{IEEEkeywords}

% ---------------------------------------------------------------
\section{Introduction}

More and more, Bangla social media users get their news through \textit{photocards}: compact, branded graphics that pack a headline, a quotation, or a claim into one shareable image. They are quick to read and quick to forward, and that is exactly what makes them useful to bad actors. Copy a trusted outlet's typography, colour scheme, and logo placement, drop a fabricated quotation onto the lookalike template, and within minutes the result is in circulation. Once the misleading message is baked into pixels, it borrows the reach of the genuine brand while leaving the brand's accountability behind.

The defenses we have today are a poor fit for this threat. Most automated misinformation systems work on text~\cite{b2,b3}, combing articles, posts, or transcripts for linguistic and stylistic anomalies. A photocard offers no such text to comb: the words are part of the raster, invisible to a tokenizer. One could, in principle, bolt optical character recognition (OCR) onto the front of a text pipeline, but OCR on stylized, low-resolution, heavily compressed graphics is unreliable, slow, and---even when it works---returns only the words while discarding the very brand-level visual cues that separate a real card from a counterfeit. For a low-resource language like Bangla, where annotated misinformation corpora are still scarce~\cite{b4}, the mismatch only widens.

Our stance is different: for this medium, authenticity is a property of the \textit{image}, not of the words printed on it. Genuine photocards from a given outlet follow a strict visual grammar---fixed typefaces, a brand palette, exact logo coordinates, standardized margins---because a template produces them. A forgery is a hand-built approximation of that template, and approximating something leaves marks: misaligned glyphs, double JPEG compression boundaries, font anti-aliasing the brand's renderer would never produce, and small drifts away from the canonical layout. A casual reader misses most of these residuals, yet statistically they are quite reliable. Picking them out is a forensics problem, a close cousin of general image-manipulation detection~\cite{b18,b19,b20,b21} but narrowed to one structured visual domain.

There is a practical payoff to framing things this way. A model that reads a photocard straight from its pixels needs no OCR, barely cares what language is printed on the card, and can be kept small enough to run at share time on ordinary hardware. That leaves two empirical questions: what architecture the task demands, and how to gauge its accuracy honestly on a heavily augmented corpus, where a careless data split can quietly inflate the numbers. We take both in turn---first showing what fails, then what works, and finally measuring how much a leakage-free protocol shifts the reported accuracy.

Our contributions are as follows.
\begin{enumerate}
    \item \textbf{A localized photocard dataset with documented provenance.} We assemble 2,047 photocards scraped exclusively from public Facebook pages---the verified pages of seventeen Bangladeshi television broadcasters on the authentic side; the pseudo-broadcaster pages, anonymous profiles, and groups that push fabricated cards on the other---of which 2,046 (969 authentic, 1,077 fabricated) survive integrity filtering. Per-page source counts are reported in full (Tables~\ref{tab:src_real} and~\ref{tab:src_fake}), alongside a reproducible preprocessing pipeline (see the Data and Code Availability statement).
    \item \textbf{A degradation-aware augmentation and integrity pipeline.} Sixteen photometric and morphological transformations grow the corpus to 32,736 images that imitate real-world social-media artefacts, after which a byte-level integrity pass guarantees that every file decodes cleanly.
    \item \textbf{A diagnostic baseline.} A fully connected ANN drops to near-random accuracy with its loss pinned at $\ln 2$, turning ``flat architectures cannot represent the relational forensic signal'' from an assertion into a measured fact.
    \item \textbf{A compact, fully normalized CNN, and a fix for a degenerate failure mode.} A four-block batch-normalized CNN with 489,954 parameters---$1{,}644\times$ fewer than the ANN---learns the task on CPU. We also pin down and correct a one-class collapse traced to an unnormalized classifier head.
    \item \textbf{Leakage-free evaluation and a measured transfer comparison.} We evaluate with stratified \emph{group} five-fold cross-validation that splits by original image, removing augmentation leakage, and we report per-class precision, recall, $F_1$, and pooled confusion matrices. Inside this protocol we pit a from-scratch CNN against one whose head is seeded from the ANN.
    \item \textbf{A strong pretrained reference.} We fine-tune an ImageNet-pretrained YOLOv8n-cls backbone over the very same leakage-free folds; at $97.36\%$ pooled accuracy it clears the compact CNNs by roughly eight points and pins down the headroom a larger, pretrained model buys over a purpose-built one.
    \item \textbf{An error-analysis and bias--variance diagnosis.} A manual analysis of the CNN's held-out errors, a human-level estimate, and an inference-mode bias--variance breakdown show the compact CNNs are under-fitting---high avoidable bias, negligible variance---which both explains the pretrained model's jump and points to bias reduction, not regularization, as the right next step.
    \item \textbf{A transparent accounting.} We show in concrete numbers that the leakage-free protocol pulls a previously rosy augmentation-level estimate ($\sim$96.8\%) down to a realistic $\sim$87--89\%, and we lay out the threats to validity that remain.
\end{enumerate}

% ---------------------------------------------------------------
\section{Related Work}

\subsection{Misinformation and fake-news detection}
Automated misinformation detection has come a long way, yet the prevailing approach is still text-centric. Data-mining surveys organize the task around linguistic cues, social context, and propagation patterns~\cite{b2,b3}, and the best systems mine article content, user networks, or both. Such methods work well when a textual signal is present, but they are blind by construction to claims that live inside an image. For Bangla the gap is especially wide: resources like BanFakeNews~\cite{b4} have begun to tackle Bangla text, yet nothing comparable targets the photocard format, which is how much of the misinformation actually travels. Media-studies research, for its part, shows experimentally that the visual side of a news item shapes framing effects on what audiences think and intend to do---sometimes independently of the accompanying text, sometimes more strongly than it~\cite{b1}---which only strengthens the case for treating the image as the thing to scrutinize. Multimodal detectors such as SpotFake~\cite{b5} do fuse text and image features, but they still assume an accompanying textual post and never put the card itself under forensic examination.

\subsection{Image forgery and manipulation detection}
In the forensics community the field has shifted firmly from handcrafted features to end-to-end learning. Convolutional networks now learn manipulation-sensitive representations straight from pixels. Bayar and Stamm constrain the first convolutional layer so that it suppresses image content and surfaces tampering residuals~\cite{b18}; Zhou et al.\ pair an RGB stream with a noise stream to localize splicing and removal~\cite{b19}; and ManTra-Net frames manipulation tracing as anomaly detection across a broad family of forgery types~\cite{b20}. Wider surveys of media forensics and deepfakes record both the progress and the fragility of these detectors once images are post-processed~\cite{b21}, and face-centric benchmarks such as FaceForensics++~\cite{b22} have standardized how synthetic-media detection is evaluated. What CNNs are good at---localized, high-frequency cues---lines up neatly with the glyph- and logo-level artefacts that give a forged photocard away.

\subsection{Architectural and training foundations}
The convolutional inductive biases we build on---local connectivity, weight sharing, translation equivariance---go back to early gradient-based recognition~\cite{b6} and were scaled up by AlexNet~\cite{b7}, VGG~\cite{b8}, and residual networks~\cite{b9}, with EfficientNet later putting the accuracy--efficiency trade-off on a principled footing through systematic scaling~\cite{b10}. Global average pooling, which we use to join the convolutional backbone to the classifier, was proposed as a parameter-free, regularizing substitute for large flatten-to-dense bottlenecks~\cite{b11}. Batch normalization~\cite{b14} steadies and speeds up training by normalizing layer activations, while dropout~\cite{b12} regularizes the dense head; both sit at the heart of the model that succeeds here. Warm-starting a new model from already-learned weights is a well-worn way to converge faster and more stably~\cite{b15}, and how well it transfers comes down to how task- and shape-compatible the donor and recipient layers are~\cite{b16}. Beyond bespoke classifiers, the single-stage detector families typified by YOLO~\cite{b23} fold feature extraction and prediction into one efficient network and ship with strong ImageNet-pretrained classification variants; we take one such variant, YOLOv8n-cls, as a compact pretrained yardstick against our from-scratch design.

\subsection{Leakage-aware evaluation}
A recurring failure mode in applied machine learning is data leakage between training and evaluation, which yields accuracy that looks impressive but does not reproduce. With a heavily augmented dataset, a naive split over the augmented files drops near-duplicate variants of one source on both sides of the partition. The usual fix is \emph{grouped} cross-validation, where every sample derived from a single source stays in the same fold. We make stratified group $k$-fold the backbone of our evaluation for exactly this reason.

% ---------------------------------------------------------------
\section{Problem Setting}
\label{sec:formulation}

Let $\mathbf{x}\in\mathbb{R}^{H\times W\times 3}$ be a photocard rendered as an RGB raster and let $y\in\{0,1\}$ be its provenance label, the two indices standing for the fabricated and authentic classes. Our goal is a classifier $f_\theta$ that sends an image to a probability over the two classes and minimizes the cross-entropy loss of Section~\ref{sec:loss} over the labelled corpus.

The premise behind pixel-level detection is easy to state. Authentic cards from a given outlet come out of a near-deterministic template, so they share a tightly constrained visual grammar; a forgery is a hand-rebuilt approximation of that same template. The discriminative evidence, then, sits in \emph{relational} cues---an edge discontinuity, a recompression block boundary, an anti-aliasing mismatch, a logo nudged a few pixels off---each one defined by how \emph{adjacent} pixels relate rather than by their isolated intensities. That is precisely the property a flatten operation throws away and a convolution keeps. Section~\ref{sec:methodology} turns the distinction into two concrete architectures, and Section~\ref{sec:results} shows it is decisive in practice.

% ---------------------------------------------------------------
\section{Dataset Creation and Preparation}

One of the central contributions of this work is the dataset itself, since no public corpus addresses the Bangla photocard format. Table~\ref{tab:dataset} lays out its composition stage by stage. Throughout, we treat the operational labels of the released corpus---\textit{Real News} for authentic cards, \textit{False News} for fabricated ones---as interchangeable with ``authentic'' and ``fabricated.''

\begin{table}[htbp]
    \centering
    \caption{Dataset Composition at Each Preparation Stage}
    \label{tab:dataset}
    \renewcommand{\arraystretch}{1.2}
    \begin{tabular}{lrrl}
        \toprule
        \textbf{Stage} & \textbf{Authentic} & \textbf{Fabricated} & \textbf{Total} \\
        \midrule
        Collected (Facebook)  & 970    & 1,077  & 2,047  \\
        After integrity pass  & 969    & 1,077  & 2,046  \\
        Post-Augmentation     & 15,504 & 17,232 & 32,736 \\
        \bottomrule
    \end{tabular}
\end{table}

\subsection{Data collection and source pages}
Every card in the corpus was scraped from Facebook, and only from Facebook. That restriction was a deliberate choice rather than a convenience: in Bangladesh the photocard is a Facebook-native format, published there by the outlets themselves and forged there by the actors who imitate them, so collecting anywhere else would have meant studying the medium outside its natural habitat. Provenance was recorded at collection time---each file preserves the handle of the page it was taken from---and in most cards the origin is also visible in the pixels, since outlets and forgers alike stamp their cards with a logo block and a footer carrying the page's name, web address, and social handles.

\subsubsection{Authentic sources}
The authentic class draws on the verified Facebook pages of seventeen national television broadcasters, listed with their per-page counts in Table~\ref{tab:src_real}. Verification is what anchors the class: a card taken from a verified page is, by the platform's own standard, a genuine product of that outlet's editorial template. The contributions are reasonably even---News24 (95 cards) and Somoy TV (92) sit at the top, ATN News (41) at the tail---so no single broadcaster supplies more than about ten percent of the class, and the learned notion of ``authentic'' is not tied to any one house style.

\begin{table}[htbp]
    \centering
    \caption{Authentic Class: Verified Broadcaster Pages\\(counts after integrity filtering)}
    \label{tab:src_real}
    \renewcommand{\arraystretch}{1.15}
    \begin{tabular}{llr}
        \toprule
        \textbf{Outlet} & \textbf{Facebook page} & \textbf{Cards} \\
        \midrule
        News24            & \texttt{news24bd\_tv}      & 95 \\
        Somoy TV          & \texttt{somoynews\_tv}     & 92 \\
        Ekushey TV        & \texttt{Ekushey24online}   & 73 \\
        Independent TV    & \texttt{itvbd24\_tv}       & 60 \\
        NTV               & \texttt{ntvdigital}        & 59 \\
        Jamuna Television & \texttt{JamunaTelevision}  & 57 \\
        Boishakhi TV      & \texttt{BoishakhiTvNews}   & 54 \\
        Banglavision      & \texttt{BanglaVisionNews}  & 52 \\
        Channel i         & \texttt{channelitv}        & 51 \\
        Maasranga TV      & \texttt{maasrangatvnews}   & 50 \\
        SA TV             & \texttt{satvbd}            & 50 \\
        Ekattor TV        & \texttt{EkattorTV}         & 50 \\
        DBC News          & \texttt{dbcnews\_tv}       & 48 \\
        RTV               & \texttt{rtvonline}         & 47 \\
        GTV               & \texttt{GTVonlineNews}     & 47 \\
        BTV               & \texttt{btv\_gov\_bd}      & 43 \\
        ATN News          & \texttt{ATNNews}           & 41 \\
        \midrule
        \textbf{Total}    &                            & \textbf{969} \\
        \bottomrule
    \end{tabular}
\end{table}

\subsubsection{Fabricated sources}
The fabricated class was collected at the source: from the pages, profiles, and groups that actually manufacture and circulate fake photocards, rather than second-hand through fact-checking articles. Table~\ref{tab:src_fake} lists them. The most striking pattern is a family of \emph{pseudo-broadcasters}---pages styled as television channels (``Gupto TV,'' ``Achuda TV,'' ``Jawra TV,'' ``Chalooo TV,'' ``Anwar TV,'' and several smaller ones) that correspond to no licensed broadcaster, yet dress every card in the full visual grammar of a TV news card: a logo block, a dated headline strip, and a footer of web and social handles. Alongside these sit three anonymous numeric profiles that together supply the single largest share of the class (408 cards) and stamp their output with the same kind of invented channel branding; a parody masthead that plays on a real newspaper's name (``The Delhi Star,'' after \textit{The Daily Star}); a handful of public groups; and six cards saved individually before their page of origin could be recorded.

\begin{table}[htbp]
    \centering
    \caption{Fabricated Class: Source Pages, Profiles, and Groups}
    \label{tab:src_fake}
    \renewcommand{\arraystretch}{1.15}
    \begin{tabular}{lr}
        \toprule
        \textbf{Source} & \textbf{Cards} \\
        \midrule
        Anonymous profiles (three numeric handles) & 408 \\
        ``Gupto TV'' (\texttt{guptobangla})        & 267 \\
        ``Achuda TV'' (\texttt{Achuda2k4})         & 170 \\
        ``Jawra TV'' (\texttt{jawratv49})          & 113 \\
        ``Chalooo TV'' (\texttt{chaloootv\_news})  & 34 \\
        ``C16 Archive'' (\texttt{C16archive})                        & 20 \\
        Public Facebook groups                     & 17 \\
        ``Anwar TV'' (\texttt{anwartvo01})         & 16 \\
        ``Somoyer Khobor'' (\texttt{somoyerkhobor24\ldots}) & 9 \\
        ``The Delhi Star'' (\texttt{thedelhistar}) & 8 \\
        ``Kamla TV'' (\texttt{kamlatvofficial})    & 4 \\
        ``Chodon TV'' (\texttt{chodontv1})         & 4 \\
        ``News Bangla'' (\texttt{newsbanglabdpage})                 & 1 \\
        Saved individually (page not recorded)     & 6 \\
        \midrule
        \textbf{Total}                             & \textbf{1,077} \\
        \bottomrule
    \end{tabular}
\end{table}

\begin{figure*}[t]
    \centering
    \begin{subfigure}[t]{0.235\textwidth}
        \includegraphics[width=\linewidth]{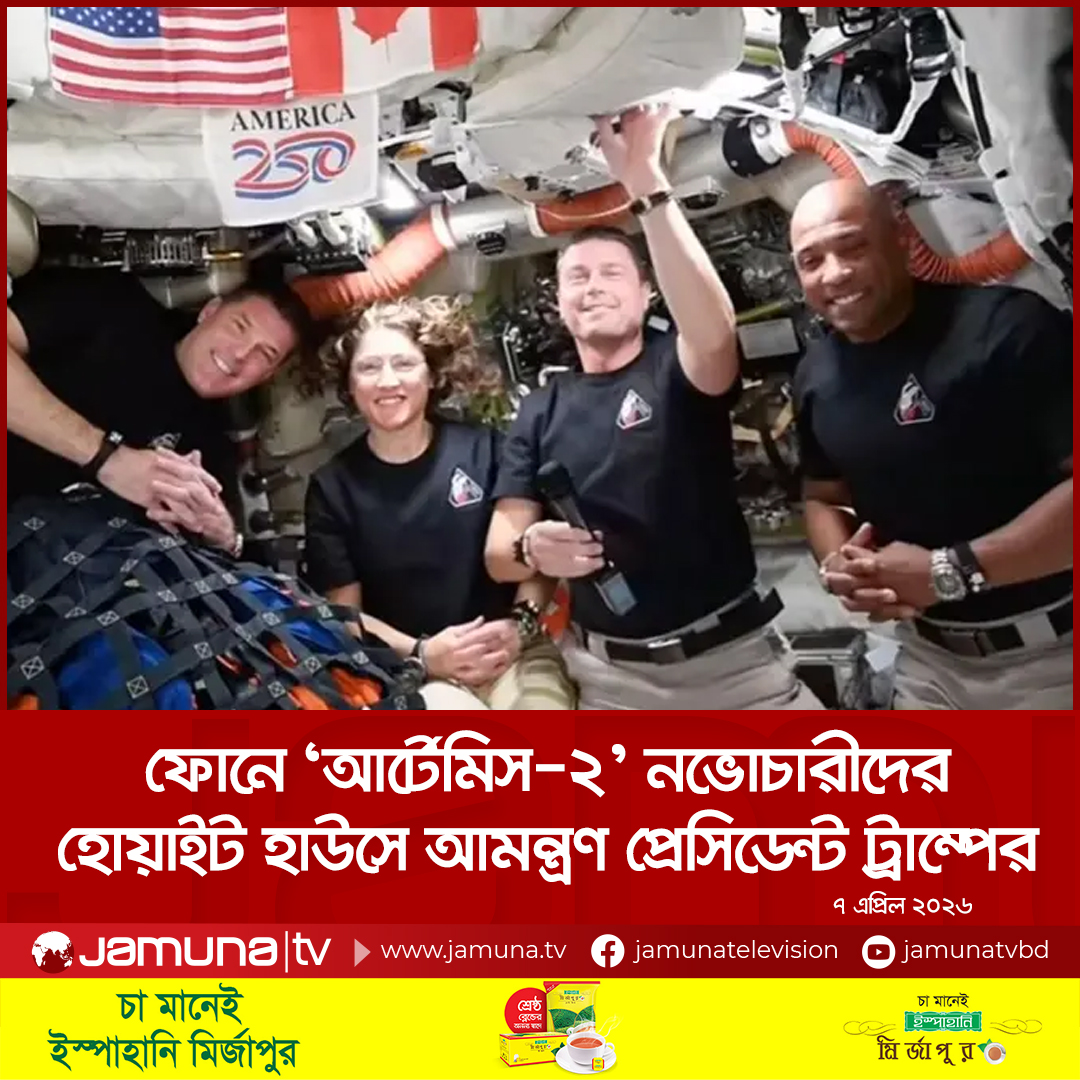}
        \caption{Authentic photocard (1)}
    \end{subfigure}\hfill
    \begin{subfigure}[t]{0.235\textwidth}
        \includegraphics[width=\linewidth]{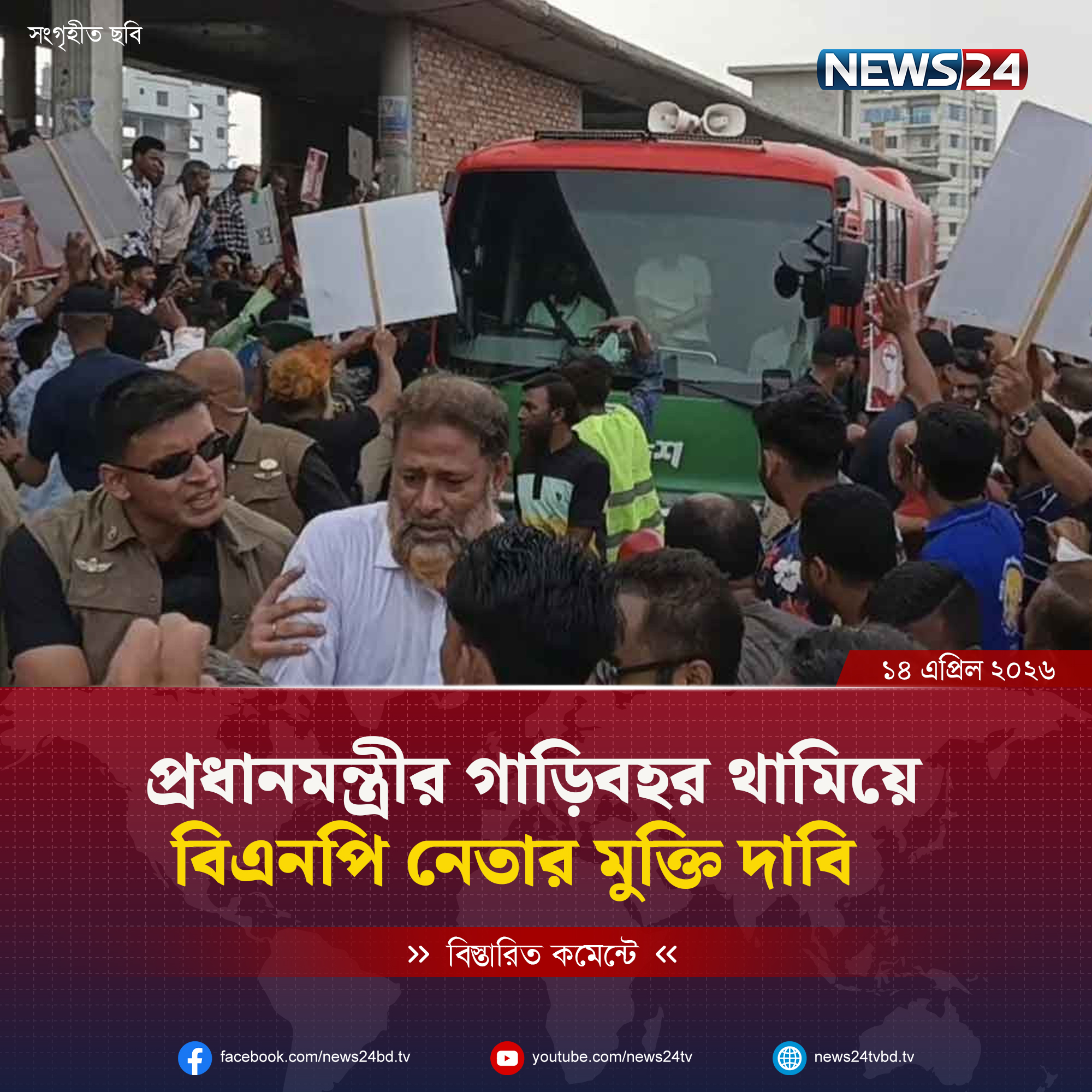}
        \caption{Authentic photocard (2)}
    \end{subfigure}\hfill
    \begin{subfigure}[t]{0.235\textwidth}
        \includegraphics[width=\linewidth]{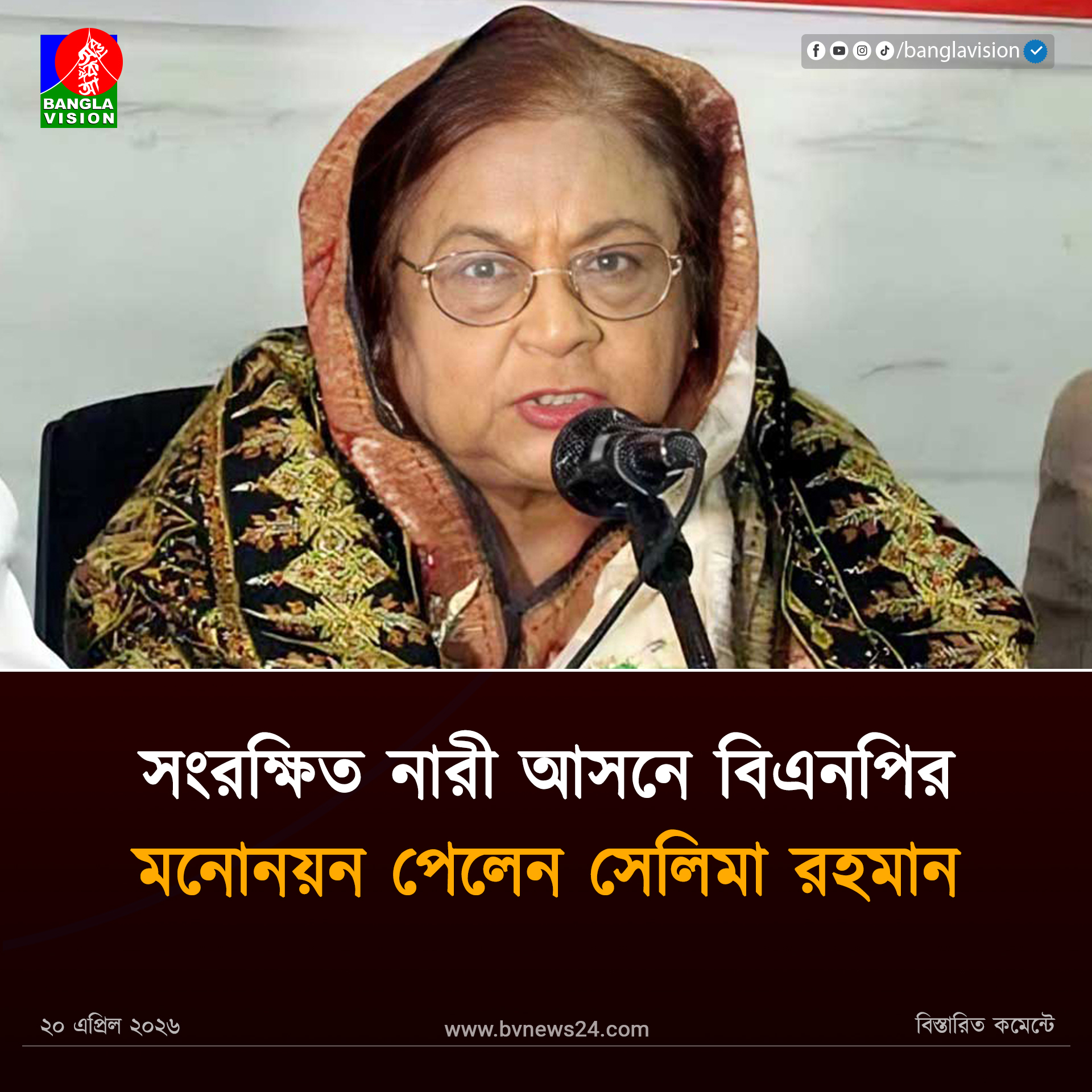}
        \caption{Authentic photocard (3)}
    \end{subfigure}\hfill
    \begin{subfigure}[t]{0.235\textwidth}
        \includegraphics[width=\linewidth]{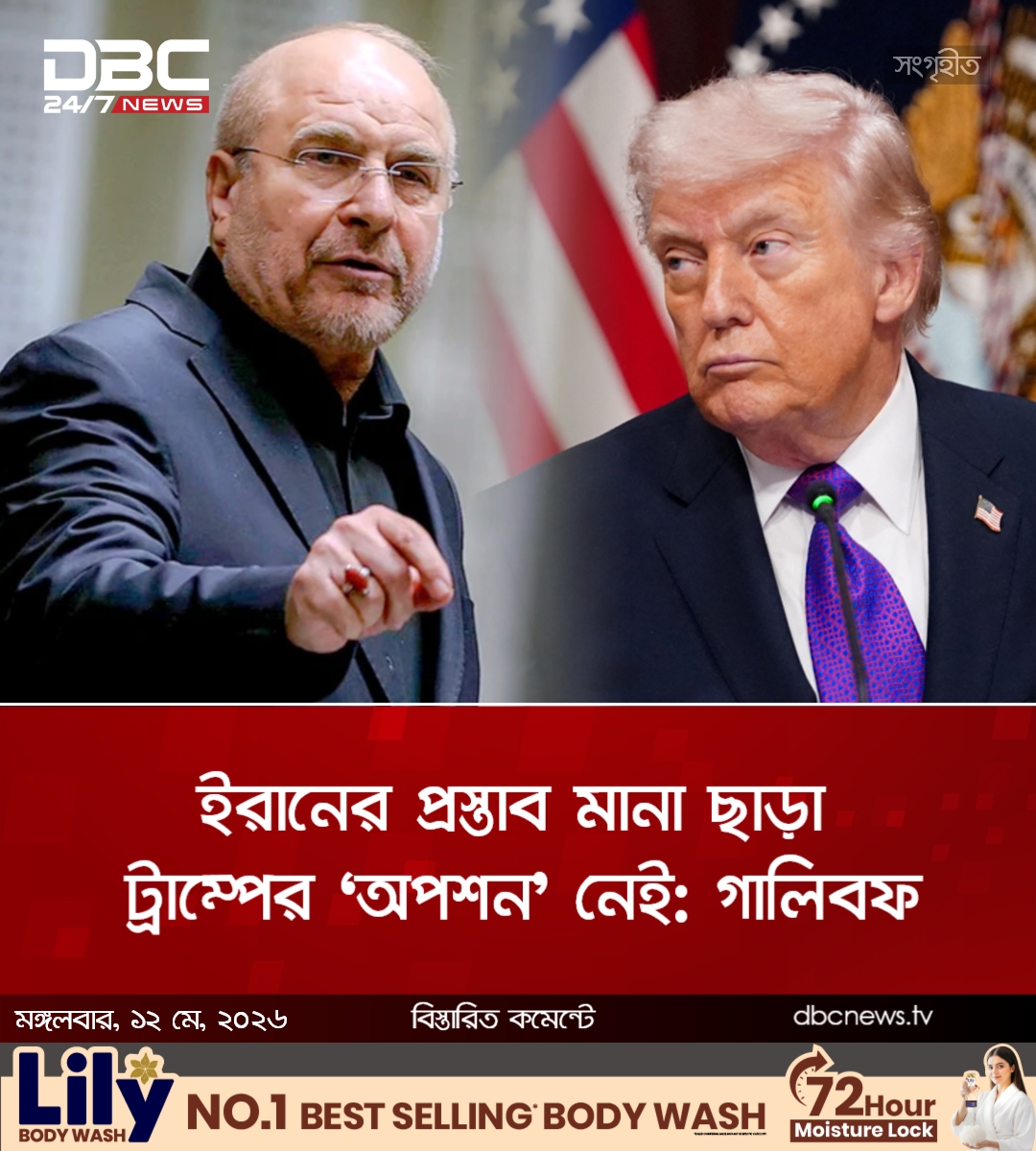}
        \caption{Authentic photocard (4)}
    \end{subfigure}

    \vspace{6pt}

    \begin{subfigure}[t]{0.235\textwidth}
        \includegraphics[width=\linewidth]{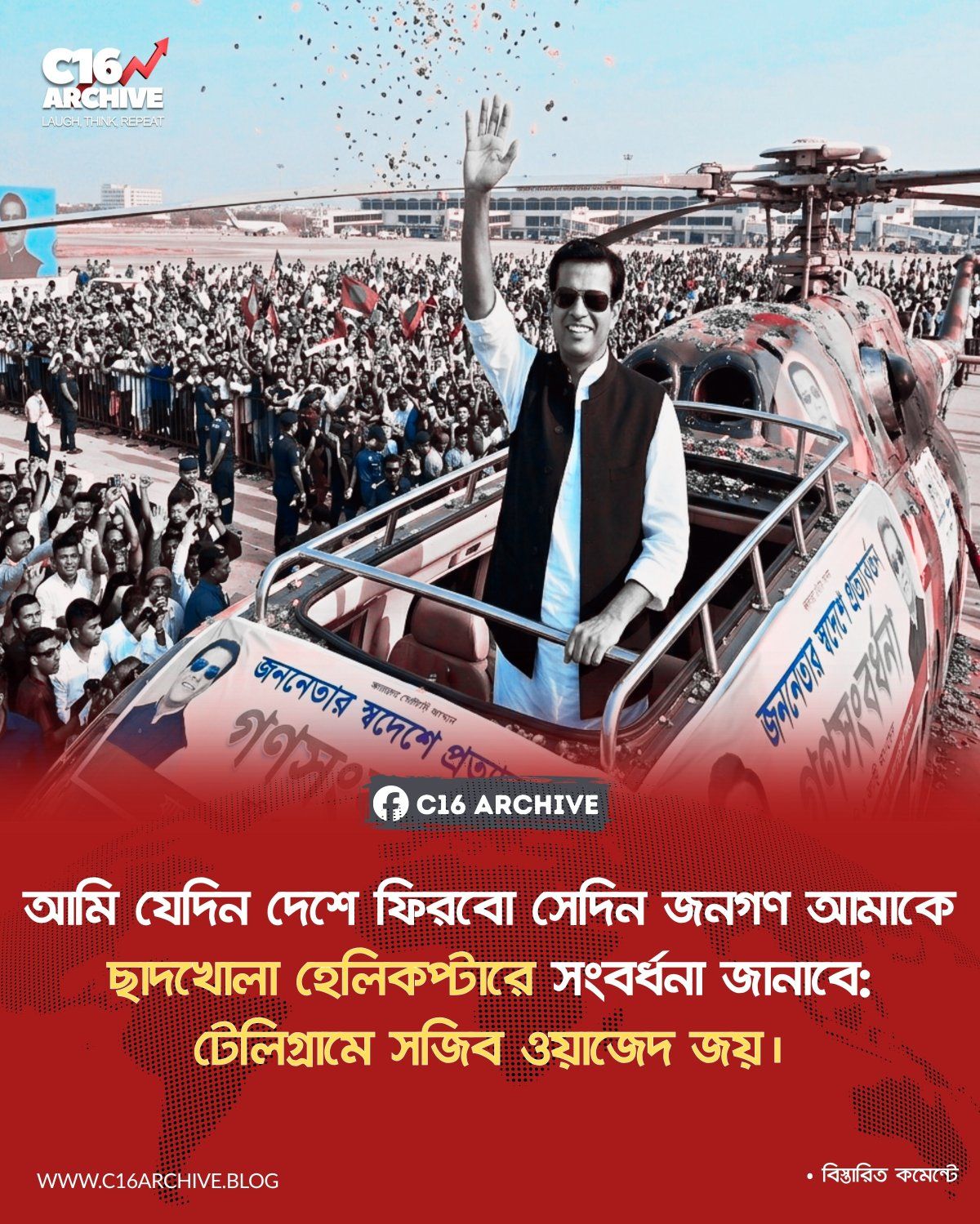}
        \caption{Fabricated photocard (1)}
    \end{subfigure}\hfill
    \begin{subfigure}[t]{0.235\textwidth}
        \includegraphics[width=\linewidth]{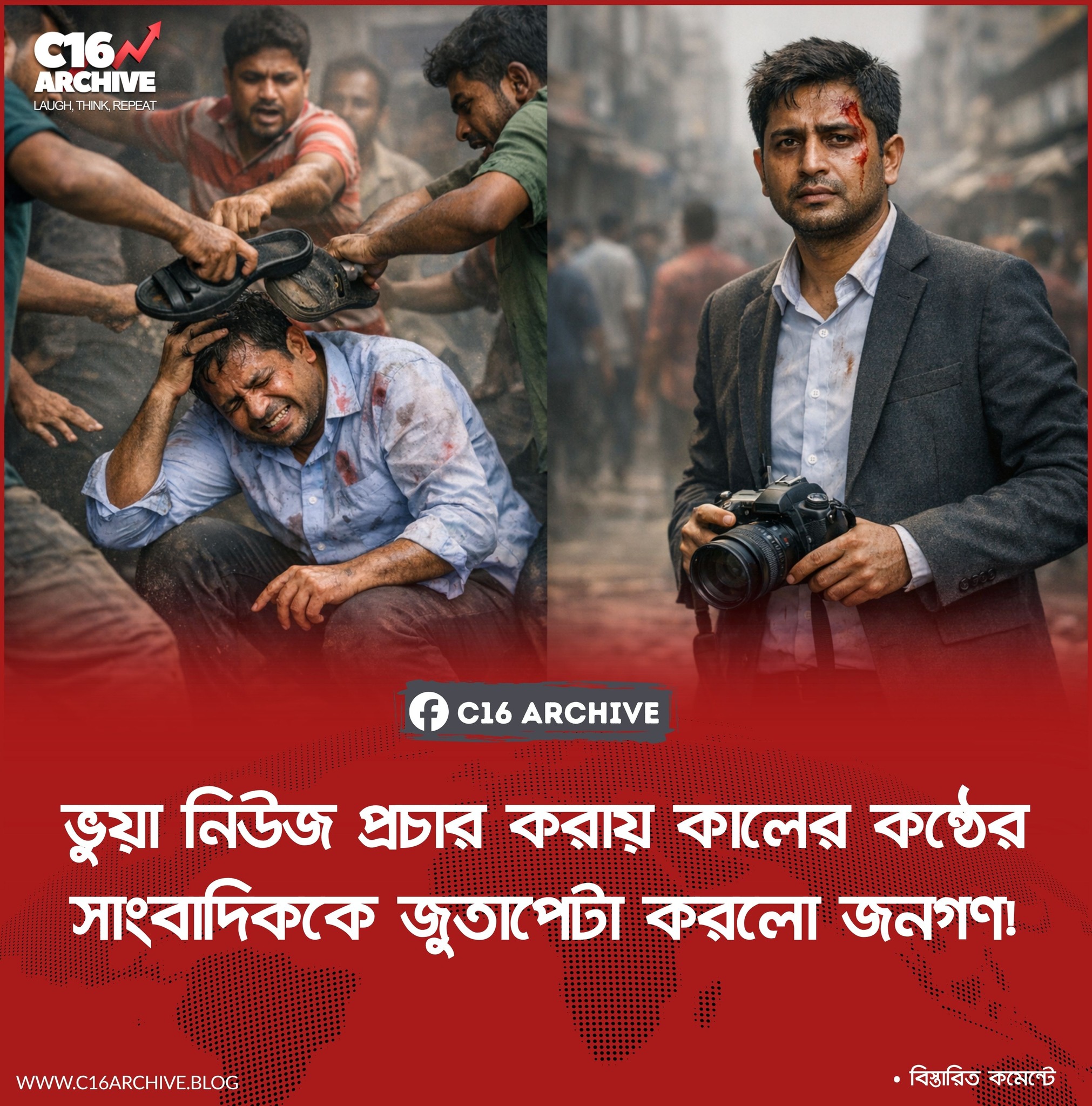}
        \caption{Fabricated photocard (2)}
    \end{subfigure}\hfill
    \begin{subfigure}[t]{0.235\textwidth}
        \includegraphics[width=\linewidth]{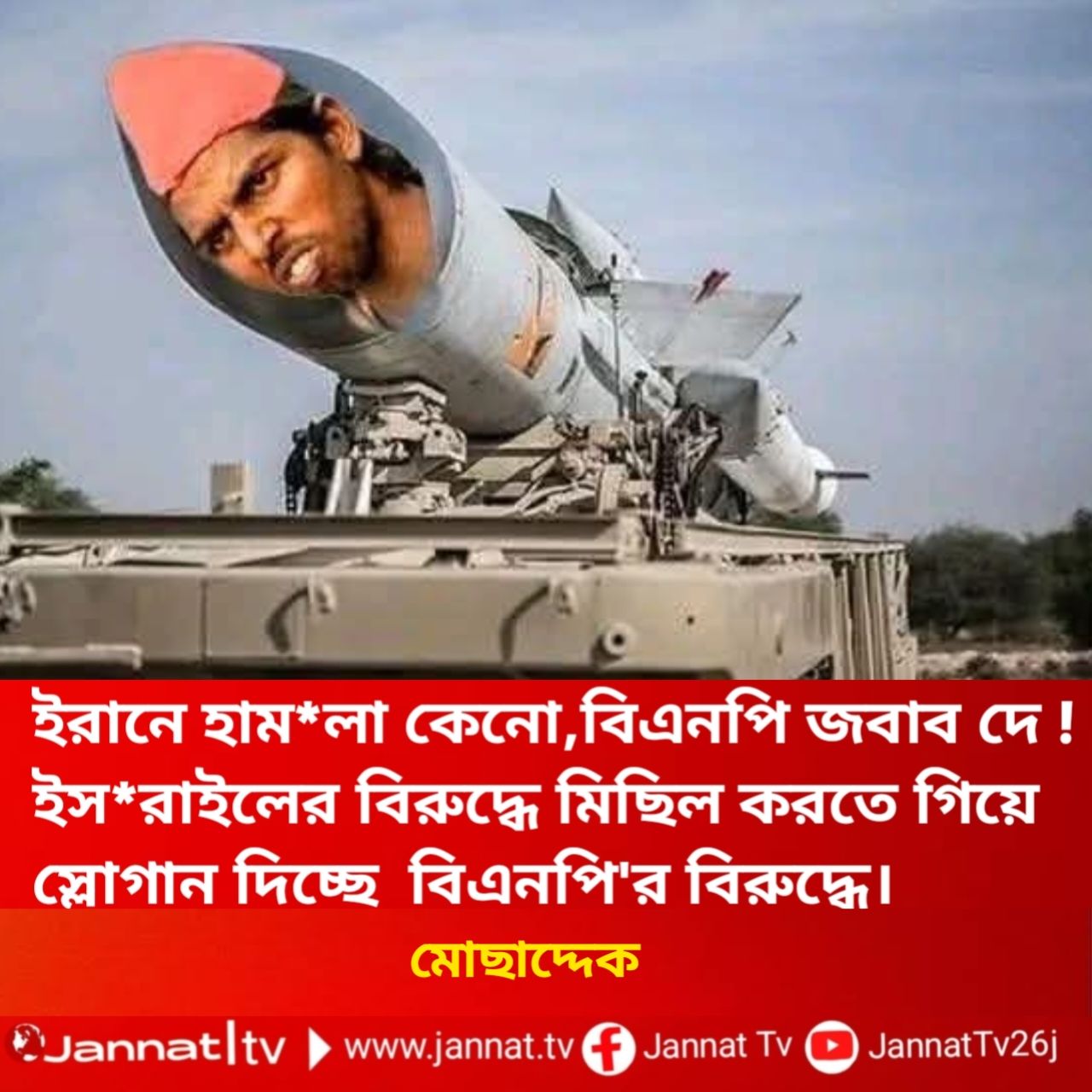}
        \caption{Fabricated photocard (3)}
    \end{subfigure}\hfill
    \begin{subfigure}[t]{0.235\textwidth}
        \includegraphics[width=\linewidth]{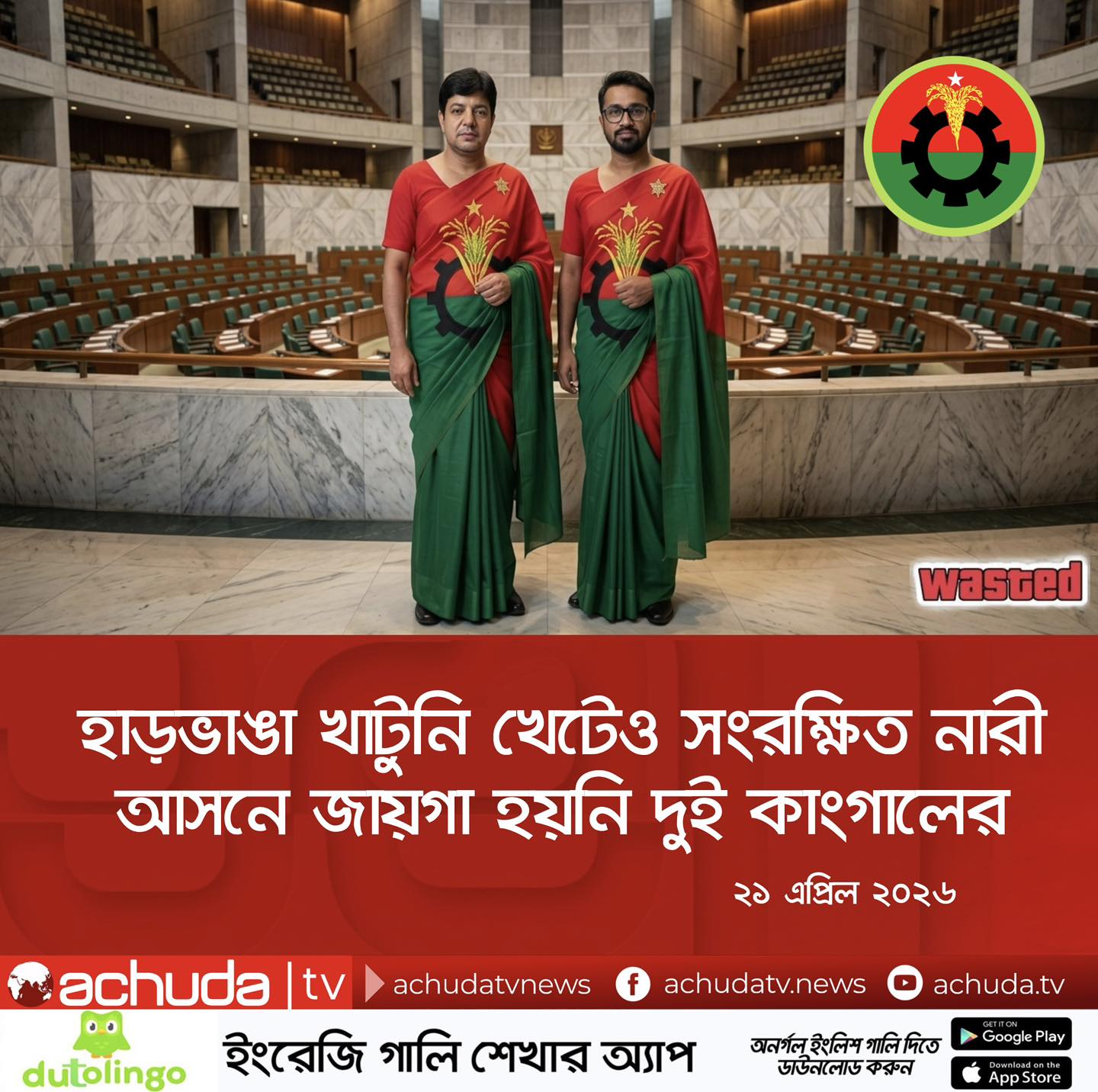}
        \caption{Fabricated photocard (4)}
    \end{subfigure}

    \caption{Representative samples. Top row: authentic photocards from verified broadcaster pages. Bottom row: fabricated photocards from the Facebook pages and profiles that circulate them. Cues like inconsistent fonts, misplaced logos, and irregular palettes are subtle to a human reader yet act as detectable digital fingerprints.}
    \label{fig:samples}
\end{figure*}

Where the authentic class is spread thin across many pages, the fabricated class is heavily concentrated: its three most prolific sources account for roughly three quarters of all fake cards (813 of 1,077). Figure~\ref{fig:samples} shows representative cards from both classes. Every image in both classes comes from publicly posted content. We return to the provenance asymmetry between the two classes---and the shortcut-learning risk it brings---explicitly in Section~\ref{sec:limitations}, since we believe it matters for reading our results correctly.

\subsection{Augmentation pipeline}
An image that has been through a few rounds of sharing is rarely pristine: every platform it passes recompresses, crops, resizes, and re-colours it. A detector trained only on clean source cards would overfit to that cleanliness and stumble in deployment. To close the domain gap---and following the broad evidence that augmentation helps image models generalize~\cite{b17}---we apply sixteen distinct morphological and photometric operations to each base image (Table~\ref{tab:augmentations}), bringing the corpus to 32,736 images after integrity filtering, roughly sixteen times the raw count, with the exact per-class totals in Table~\ref{tab:dataset}. The operations cover geometric distortion (rotation, flipping, scaling, cropping, warping), photometric perturbation (brightness, contrast, additive noise, light and heavy blur), and a handful of compound pipelines that stack three or four operations to reproduce the piled-up degradation of multi-platform re-sharing.

\begin{table}[htbp]
    \centering
    \caption{Augmentation Transformations Applied to Each Base Image}
    \label{tab:augmentations}
    \renewcommand{\arraystretch}{1.15}
    \begin{tabular}{ll}
        \toprule
        \multicolumn{2}{c}{\textbf{Transformations}} \\
        \midrule
        Rotate                     & Flip \\
        Scale                      & Crop \\
        Warp                       & Brightness \\
        Contrast                   & Heavy Blur \\
        Light Blur                 & Noise \\
        Crop--Tilt--Blur--Contrast & Noisy--Bright--Crop \\
        Warp--Blur--Noise          & Scale--Contrast--Noise \\
        Crop--Flip--Bright--Blur   & Heavy Distortion \\
        \bottomrule
    \end{tabular}
\end{table}

The resulting class split (15,504 authentic against 17,232 fabricated) is close to balanced, a ratio of about $1{:}1.11$. That balance helps in two ways: it keeps accuracy a meaningful summary statistic, and it stops a majority-class prior from taking over the decision boundary during optimization. To keep the mild leftover imbalance from biasing the from-scratch model, we also apply balanced class weights during training (Section~\ref{sec:methodology}).

\subsection{Data integrity and deep cleaning}
A single undecodable file can kill a TensorFlow input pipeline mid-epoch, so before any training we swept every directory with an integrity pass. The script decodes each file end-to-end through TensorFlow's image I/O path---the same path the training loader uses---and deletes anything that throws a decoding error or shows byte-level corruption. This way the corpus that reaches the optimizer is structurally sound, and no silent read failure can contaminate a batch. On the present corpus the pass removed exactly one file---a Somoy TV card whose bytes no longer decoded cleanly---which is what trims the authentic class from the 970 cards collected to the 969 that enter augmentation (Table~\ref{tab:dataset}). The pass is deterministic and rerunnable, so the dataset stays reproducible from one machine to the next.

% ---------------------------------------------------------------
\section{Methodology}
\label{sec:methodology}

We have organized the methodology as an argument rather than a catalogue. We start with the simplest plausible model, work out exactly why it cannot succeed, and let that failure shape the model that does. From there we explain how the working model is trained and---just as important---how it is evaluated without leakage.

\subsection{Baseline: Artificial Neural Network}
To sanity-check the data pipeline and set a reference point, we began with a fully connected ANN. Images were loaded and resized to $1024\times1024$ pixels to keep fine detail, the corpus was shuffled and split into training and validation subsets, and each step drew batches of 128 images. Table~\ref{tab:ann_arch} gives the architecture; we compiled the model with the Adam optimizer and sparse categorical cross-entropy and trained it for 50 epochs.

\begin{table}[htbp]
    \centering
    \caption{Baseline ANN Architecture}
    \label{tab:ann_arch}
    \renewcommand{\arraystretch}{1.2}
    \begin{tabular}{llr}
        \toprule
        \textbf{Layer} & \textbf{Configuration} & \textbf{Output Shape} \\
        \midrule
        Rescaling & $\times\,\frac{1}{255}$ & $1024\times1024\times3$ \\
        Flatten   & ---                     & $3{,}145{,}728$ \\
        Dense     & 256 units, ReLU         & $256$ \\
        Dropout   & rate $= 0.30$           & $256$ \\
        Dense     & 128 units, ReLU         & $128$ \\
        Dense     & 2 units, Softmax        & $2$ \\
        \bottomrule
    \end{tabular}
\end{table}

\subsection{Why a flattened model is structurally inadequate}
Everything about the ANN's behaviour follows from its first operation. Flattening turns the $1024\times1024\times3$ input into a vector of $3{,}145{,}728$ independent coordinates, and from that point on the spatial neighbourhood of any two pixels is unrecoverable. The first dense layer then has to learn a separate weight for every pixel--unit pair---around $805$ million parameters---before it can express even one forensic pattern. The catch is that the forensic signal is \emph{relational}: an edge discontinuity, a compression block boundary, or an anti-aliasing mismatch is defined by how neighbouring pixels relate, not by their isolated intensities. A representation that has thrown away adjacency cannot encode such features efficiently even in principle, and gradient descent over a $3.1$-million-dimensional input carries no inductive bias that would let it rediscover them. Dropout~\cite{b12} cannot save a representation that was impoverished from the start. The empirical consequence---a collapse to the random-classifier floor---appears in Section~\ref{sec:results}.

\subsection{CNN architecture}
The corrective architecture holds on to spatial structure all the way through feature extraction. We dropped the input resolution to $256\times256$, which keeps activation and gradient memory modest while the hierarchical receptive fields of the convolutional stack make up for the lower resolution. The network runs four convolutional blocks, a Global Average Pooling (GAP) layer, and a two-layer dense classifier (Table~\ref{tab:cnn_arch}). Each block is a $3\times3$ padded convolution, batch normalization~\cite{b14}, a ReLU nonlinearity, and $2\times2$ max pooling, with the filter count doubling block to block ($32\!\to\!64\!\to\!128\!\to\!256$). Weight sharing makes each convolution translation-equivariant, so a kernel fires on a forgery artefact wherever it lands on the card, while max pooling halves the spatial resolution and widens the effective receptive field. GAP then reduces each of the 256 final channels to its spatial mean, yielding a fixed 256-dimensional descriptor with no parameters at all; this sidesteps, by design, the flatten-to-dense bottleneck that dominated the ANN's parameter count and doubles as a structural regularizer~\cite{b11}.

\begin{table}[htbp]
    \centering
    \caption{CNN Architecture with Output Shapes}
    \label{tab:cnn_arch}
    \renewcommand{\arraystretch}{1.2}
    \setlength{\tabcolsep}{4pt}
    \begin{tabular}{lll}
        \toprule
        \textbf{Layer} & \textbf{Configuration} & \textbf{Output Shape} \\
        \midrule
        Rescaling          & $\times\,\frac{1}{255}$                 & $256\times256\times3$  \\
        Conv+BN+ReLU+Pool  & 32 filters, $3\times3$                  & $128\times128\times32$ \\
        Conv+BN+ReLU+Pool  & 64 filters, $3\times3$                  & $64\times64\times64$   \\
        Conv+BN+ReLU+Pool  & 128 filters, $3\times3$                 & $32\times32\times128$  \\
        Conv+BN+ReLU+Pool  & 256 filters, $3\times3$                 & $16\times16\times256$  \\
        GlobalAvgPool2D    & ---                                     & $256$                  \\
        Dense+BN+ReLU      & 256 units (\texttt{dense1})             & $256$                  \\
        Dropout            & rate $= 0.30$                           & $256$                  \\
        Dense+BN+ReLU      & 128 units (\texttt{dense2})             & $128$                  \\
        Dense              & 2 units, Softmax (\texttt{output})      & $2$                    \\
        \bottomrule
    \end{tabular}
\end{table}

\subsubsection{Normalizing the classifier head fixes a one-class collapse}
An early version of the model batch-normalized only the convolutional stack and left the dense head untouched. In that configuration training kept falling into a degenerate solution: every validation image was assigned to one class, and validation accuracy froze at the majority-class prior. The culprit is a head whose pre-activations are free to drift in scale---paired with an aggressive learning rate, the optimizer pushes the softmax toward a constant output that minimizes the loss trivially. Three changes together make training both stable and discriminative: (i) batch-normalizing both dense layers of the head (Table~\ref{tab:cnn_arch}), (ii) clipping the global gradient norm to $1.0$, and (iii) lowering the learning rate (Section~\ref{sec:setup}). With the fully normalized head reported here, training never once collapsed across any fold of either CNN variant.

\subsubsection{Parameter budget}
The convolutional design keeps the model small. Table~\ref{tab:params} sets out the budget; the convolutional and dense weight matrices carry no bias, because each is immediately followed by batch normalization, which supplies its own shift. All told, 489,954 parameters take up $1.87$\,MB in single precision, against roughly $805$ million parameters ($\sim$3.0\,GB) for the ANN---a factor of $805{,}339{,}778/489{,}954\approx1{,}644\times$, almost all of it from swapping the flatten-to-dense bottleneck for convolutions and GAP. Of the total, just 1,728 parameters (the batch-normalization running statistics) are non-trainable.

\begin{table}[htbp]
    \centering
    \caption{CNN Parameter Budget}
    \label{tab:params}
    \renewcommand{\arraystretch}{1.2}
    \begin{tabular}{lrr}
        \toprule
        \textbf{Component} & \textbf{Output Shape} & \textbf{Parameters} \\
        \midrule
        Conv1 (32)          & $128\times128\times32$ & 864 \\
        Conv2 (64)          & $64\times64\times64$   & 18,432 \\
        Conv3 (128)         & $32\times32\times128$  & 73,728 \\
        Conv4 (256)         & $16\times16\times256$  & 294,912 \\
        Batch norm (all 6)  & ---                    & 3,456 \\
        GlobalAvgPool2D     & $256$                  & 0 \\
        Dense1 (256)        & $256$                  & 65,536 \\
        Dense2 (128)        & $128$                  & 32,768 \\
        Output (2)          & $2$                    & 258 \\
        \midrule
        \textbf{Total}      & ---                    & \textbf{489,954} \\
        \bottomrule
    \end{tabular}
\end{table}

\subsection{Two model variants: from-scratch and ANN-transfer}
We look at two instances of this architecture, differing only in how the classifier head is initialized and trained.

\paragraph{Pure CNN (from scratch)} Every weight is randomly initialized and the whole network is trained jointly for ten epochs at one learning rate, with balanced class weights, a reduce-on-plateau learning-rate schedule, and early stopping that restores the best weights.

\paragraph{Transfer CNN (ANN head)} The two shape-compatible head layers are seeded from the ANN at epoch 50: the CNN's \texttt{dense2} ($256\to128$) inherits the ANN's Dense(128) weights, and the CNN's \texttt{output} ($128\to2$) inherits the ANN's Dense(2) weights. The ANN's first dense layer maps a $3{,}145{,}728$-dimensional flattened vector to 256 units, whereas the CNN's \texttt{dense1} maps a 256-dimensional GAP vector to 256 units; the two are geometrically incompatible and cannot be transferred, so \texttt{dense1} starts from a random initialization (Table~\ref{tab:transfer}). Training then runs in two phases (Algorithm~\ref{alg:training}): Phase~1 freezes the transferred head while the backbone and \texttt{dense1} warm up, and Phase~2 fine-tunes all layers at a learning rate ten times lower. Since the two variants share an architecture and an evaluation protocol, comparing them is a controlled---if not perfectly isolated---measurement of what the ANN warm-start buys; they also differ in training schedule (single- versus two-phase, plus class weighting), a point we flag rather than bury.

\begin{table}[htbp]
    \centering
    \caption{Layer-wise Transfer from ANN to CNN Classifier Head}
    \label{tab:transfer}
    \renewcommand{\arraystretch}{1.2}
    \begin{tabular}{p{1.7cm}p{1.7cm}p{3.7cm}}
        \toprule
        \textbf{Layer} & \textbf{Transferred} & \textbf{Justification} \\
        \midrule
        Dense(256) & No  & Input dims incompatible: ANN 3,145,728 vs.\ CNN 256 (GAP). \\
        Dense(128) & Yes & Shape $256\to128$ identical in both models. \\
        Dense(2)   & Yes & Shape $128\to2$ identical in both models. \\
        \bottomrule
    \end{tabular}
\end{table}

\subsection{Loss and optimization}
\label{sec:loss}
The output layer applies a softmax, and we minimize the sparse categorical cross-entropy of the true class, $\mathcal{L}=-\log \hat{p}_{y}$, averaged over each mini-batch. On a balanced binary problem, a classifier that ignores its input settles at $\hat{p}_y=\tfrac12$ and so at a loss of $\ln 2\approx0.693$; this is the diagnostic floor we lean on when reading the ANN. Adam~\cite{b13} updates the parameters with the global gradient norm clipped at $1.0$. For the from-scratch CNN we additionally weight each class inversely to its frequency,
\begin{equation}
w_c=\frac{N}{C\,n_c},
\end{equation}
where $N$ is the number of training samples, $C=2$ the number of classes, and $n_c$ the count of class $c$; this keeps a constant-class prediction from minimizing the loss under the mild $1{:}1.11$ imbalance.

\begin{algorithm}[t]
\caption{Two-Phase Transfer-Initialized CNN Training}
\label{alg:training}
\begin{algorithmic}[1]
\Require Training fold $\mathcal{D}$; ANN checkpoint $\theta^{\text{ANN}}$; phase-1 length $T_1{=}5$; total epochs $T{=}10$
\State Build CNN with head layers \texttt{dense2}, \texttt{output} shaped as in the ANN
\State $\theta_{\texttt{dense2}}\gets\theta^{\text{ANN}}_{\mathrm{Dense}(256\to128)}$;\;\; $\theta_{\texttt{output}}\gets\theta^{\text{ANN}}_{\mathrm{Dense}(128\to2)}$
\State Freeze \texttt{dense2}, \texttt{output}; compile Adam with $\eta=10^{-4}$, clipnorm $1.0$
\For{$t=1$ to $T_1$} \Comment{Phase 1: backbone warm-up, head frozen}
  \State update conv.\ blocks and \texttt{dense1} on $\mathcal{D}$
\EndFor
\State Unfreeze all layers; \textbf{recompile} fresh Adam with $\eta=10^{-5}$
\For{$t=T_1{+}1$ to $T$} \Comment{Phase 2: full fine-tuning}
  \State update all parameters on $\mathcal{D}$
\EndFor
\State \Return $f_\theta$
\end{algorithmic}
\end{algorithm}

\subsection{Strong pretrained baseline: YOLOv8-cls}
\label{sec:yolo}
The compact CNN is small by design and trained from nothing, which leaves an obvious question hanging: how much accuracy would a modern, pretrained vision backbone add on this task? To answer it---and to supply the positive pretrained baseline our earlier evaluation lacked---we fine-tune \mbox{YOLOv8n-cls}, the classification variant of the YOLOv8 family~\cite{b23}. It couples a CSP-Darknet convolutional backbone with a lightweight classification head and, at roughly $1.44$ million parameters and $3.4$ GFLOPs at $256\times256$, is still compact: about $2.9\times$ the parameter count of our CNN, yet three orders of magnitude below the ANN.

Rather than start from noise, we initialize the network from weights pretrained on ImageNet; $156$ of its $158$ layers carry over, and only the final classification layer is reset for the two photocard classes. The important point is that YOLOv8 sees \emph{exactly} the same data and folds as the CNNs: we reuse the leakage-free group five-fold assignment of Section~\ref{sec:groupcv} (seed $123$) and, for each fold, build an Ultralytics classification directory whose \texttt{train/} and \texttt{val/} subtrees are symbolic links back to the augmented images---never copies---so every model in this paper trains and validates on an identical partition. Training runs on the Ultralytics~8.4.75 / PyTorch~2.12 stack with its automatic optimizer selection, which picks AdamW (learning rate $1.67\times10^{-3}$, momentum $0.9$, weight decay $5\times10^{-4}$), a batch of $64$, ten epochs, and CPU execution. We keep the framework's default light train-time augmentation (random horizontal flip, HSV jitter, RandAugment, and random erasing), which stacks on top of our offline sixteen-transformation corpus.

Once a fold finishes training, we score its best checkpoint on the held-out fold and build the confusion matrix with the \emph{same} pure-numpy routine used for the CNNs, so every per-class metric and the pooled cross-validated matrix are computed and aggregated identically across all models. Because YOLOv8 differs from the CNNs in more than architecture---it is pretrained, optimized differently, and augmented further---we read it as a strong reference point rather than an architecture-controlled comparison, and we come back to this confound in Section~\ref{sec:limitations}.

\subsection{Leakage-free stratified group cross-validation}
\label{sec:groupcv}
Since every original image spawns sixteen augmented variants, a naive split over the augmented files would scatter near-duplicate variants of the same source card across both training and validation, leaking information and inflating accuracy. We rule this out by cross-validating over \emph{groups}: each augmented file is traced back to its original image (matched by filename stem against the raw corpus), and the original-image group becomes the unit we partition. On top of that we build a stratified group five-fold assignment (Algorithm~\ref{alg:groupcv}): within each class, the original-image groups are shuffled and dealt round-robin across the five folds, so each fold gets a balanced share of every class and---this is the crucial part---all sixteen augmentations of any original land together in a single fold. Every validation fold therefore contains only originals the model never trained on. The 32,736 images resolve into 2,046 groups, giving folds of 6,528--6,560 images apiece (Table~\ref{tab:foldsplit}).

\begin{algorithm}[t]
\caption{Stratified Group $k$-Fold Assignment}
\label{alg:groupcv}
\begin{algorithmic}[1]
\Require Samples with labels; original-image group id per sample; folds $k{=}5$; seed
\State Map each augmented file to its original-image group (each group is single-class)
\For{each class $c$}
  \State Collect the groups of class $c$; shuffle them with the seed
  \State Assign the $i$-th shuffled group to fold $(i \bmod k)$
\EndFor
\State \Return fold index for every sample (inherited from its group)
\end{algorithmic}
\end{algorithm}

\begin{table}[htbp]
    \centering
    \caption{Group Five-Fold Partition (by original image)}
    \label{tab:foldsplit}
    \renewcommand{\arraystretch}{1.2}
    \begin{tabular}{crrrr}
        \toprule
        \textbf{Fold} & \textbf{Groups} & \textbf{Images} & \textbf{Fabricated} & \textbf{Authentic} \\
        \midrule
        0 & 410 & 6,560 & 3,456 & 3,104 \\
        1 & 410 & 6,560 & 3,456 & 3,104 \\
        2 & 409 & 6,544 & 3,440 & 3,104 \\
        3 & 409 & 6,544 & 3,440 & 3,104 \\
        4 & 408 & 6,528 & 3,440 & 3,088 \\
        \bottomrule
    \end{tabular}
\end{table}

Each fold takes a turn as the held-out validation set while the other four train the model; the per-fold confusion matrices are then summed into one pooled confusion matrix in which every sample is held out exactly once, and we report per-class metrics both as the mean$\,\pm\,$standard deviation across folds and as pooled values. The whole study ran on CPU; getting through it without an accelerator is itself a finding, since it shows the approach does not hinge on GPU hardware.

% ---------------------------------------------------------------
\section{Experimental Setup}
\label{sec:setup}

The ANN and CNNs were built in TensorFlow~2.21 (Keras~3.14) on Python~3.12; the YOLOv8-cls baseline runs on the Ultralytics~8.4.75 / PyTorch~2.12 stack. Every model was trained on the same CPU workstation (12 logical cores, 47\,GB RAM, no GPU). Pixel values are rescaled to $[0,1]$ inside the Keras networks (Ultralytics applies its own normalization), and a fixed seed (123) drives both the augmentation grouping and the fold assignment, so all models share identical folds. Table~\ref{tab:config} gathers the full configuration.

\begin{table}[htbp]
    \centering
    \caption{Experimental Configuration}
    \label{tab:config}
    \renewcommand{\arraystretch}{1.2}
    \footnotesize
    \setlength{\tabcolsep}{3pt}
    \begin{tabular}{>{\raggedright\arraybackslash}p{1.4cm}>{\raggedright\arraybackslash}p{1.1cm}>{\raggedright\arraybackslash}p{2.5cm}>{\raggedright\arraybackslash}p{1.7cm}}
        \toprule
        \textbf{Setting} & \textbf{ANN} & \textbf{CNN (pure / transfer)} & \textbf{YOLOv8-cls} \\
        \midrule
        Input res.       & $1024^2$ & $256^2$ & $256^2$ \\
        Batch size       & 128 & 128 & 64 \\
        Optimizer        & Adam & Adam (clipnorm $1.0$) & AdamW (auto) \\
        Learning rate    & $1\!\times\!10^{-3}$ & $10^{-4}$ \,/\, $10^{-4}$(P1), $10^{-5}$(P2) & $1.67\!\times\!10^{-3}$ \\
        Loss             & Sparse CE & Sparse CE & Cross-entropy \\
        Normalization    & None & BatchNorm (conv + head) & BatchNorm (backbone) \\
        Dropout          & 0.30 & 0.30 & --- \\
        Class weights    & None & Balanced / None & None \\
        Epochs           & 50 & 10 \,/\, 10 (5+5) & 10 \\
        Init.            & Random & Random / Transferred+random & ImageNet \\
        Evaluation       & Single split & Group 5-fold (seed 123) & Group 5-fold (seed 123) \\
        Hardware         & \multicolumn{3}{l}{12-core CPU, 47\,GB RAM, no GPU} \\
        Software         & \multicolumn{3}{p{5.6cm}}{TensorFlow 2.21 / Keras 3.14 (ANN, CNN); Ultralytics 8.4.75 / PyTorch 2.12 (YOLOv8); Python 3.12} \\
        \bottomrule
    \end{tabular}
\end{table}

\subsection{Evaluation metrics}
For the two CNN variants and the YOLOv8-cls baseline we report---per fold and pooled across folds---accuracy together with per-class precision, recall, and $F_1$, and we include pooled confusion matrices in both raw-count and row-normalized form. Because the corpus is near-balanced, accuracy is a faithful summary rather than an artefact of class skew, and the per-class breakdown separates the two error types---wrongly flagging an authentic outlet (false positive) versus letting a forgery through (false negative)---whose costs diverge in deployment. For the ANN, the $\ln 2$ loss floor is an unambiguous reference for diagnosing the degenerate solution.

% ---------------------------------------------------------------
\section{Experimental Results and Analysis}
\label{sec:results}

\subsection{Baseline ANN}
We optimized the baseline ANN for 50 epochs, keeping a checkpoint every 10; the epoch-50 checkpoint later seeds the transfer CNN head. Figures~\ref{fig:accuracy} and~\ref{fig:loss} plot the 25-epoch window in which the whole learning dynamic plays out; from epoch 3 on the curves are flat to within a fraction of a point, and training simply carried on at that plateau to produce the transfer checkpoint.

\begin{figure}[htbp]
    \centering
    \begin{tikzpicture}
        \begin{axis}[
            width=\linewidth,
            height=5.2cm,
            xlabel={Epoch},
            ylabel={Accuracy},
            grid=both,
            grid style={line width=0.3pt, draw=gray!30},
            major grid style={line width=0.5pt, draw=gray!50},
            ymin=0.45, ymax=0.80,
            xtick={1,5,10,15,20,25},
            title={ANN --- Training and Validation Accuracy},
            legend style={at={(0.97,0.05)}, anchor=south east, font=\small},
            title style={font=\small\bfseries},
            label style={font=\small},
            tick label style={font=\footnotesize}
        ]
        \addplot[blue, thick, mark=square*, mark size=1.5pt] coordinates {
            (1,0.5756) (2,0.5823) (3,0.5279) (4,0.5279) (5,0.5279)
            (6,0.5292) (7,0.5277) (8,0.5279) (9,0.5279) (10,0.5277)
            (11,0.5279) (12,0.5281) (13,0.5307) (14,0.5313) (15,0.5275)
            (16,0.5275) (17,0.5275) (18,0.5275) (19,0.5275) (20,0.5275)
            (21,0.5275) (22,0.5275) (23,0.5274) (24,0.5275) (25,0.5271)
        };
        \addplot[red, thick, mark=triangle*, mark size=1.5pt, dashed] coordinates {
            (1,0.7235) (2,0.5224) (3,0.5224) (4,0.5224) (5,0.5224)
            (6,0.5222) (7,0.5222) (8,0.5222) (9,0.5219) (10,0.5219)
            (11,0.5219) (12,0.5296) (13,0.5222) (14,0.5221) (15,0.5221)
            (16,0.5221) (17,0.5221) (18,0.5221) (19,0.5221) (20,0.5221)
            (21,0.5221) (22,0.5221) (23,0.5221) (24,0.5221) (25,0.5222)
        };
        \legend{Training, Validation}
        \end{axis}
    \end{tikzpicture}
    \caption{Baseline ANN accuracy over the resolved 25-epoch window. Both curves flatten out near 52\%, in line with near-random binary classification; the brief $72.4\%$ validation reading at epoch 1 is an unstable first step under dropout and does not hold.}
    \label{fig:accuracy}

    \vspace{6pt}

    \begin{tikzpicture}
        \begin{axis}[
            width=\linewidth,
            height=5.2cm,
            xlabel={Epoch},
            ylabel={Loss},
            grid=both,
            grid style={line width=0.3pt, draw=gray!30},
            major grid style={line width=0.5pt, draw=gray!50},
            ymin=0.55, ymax=0.78,
            xtick={1,5,10,15,20,25},
            title={ANN --- Training and Validation Loss},
            legend style={at={(0.97,0.95)}, anchor=north east, font=\small},
            title style={font=\small\bfseries},
            label style={font=\small},
            tick label style={font=\footnotesize}
        ]
        \addplot[blue, thick, mark=square*, mark size=1.5pt] coordinates {
            (1,0.6778) (2,0.6664) (3,0.6916) (4,0.6912) (5,0.691)
            (6,0.6911) (7,0.6915) (8,0.6914) (9,0.6916) (10,0.6915)
            (11,0.6914) (12,0.6913) (13,0.6909) (14,0.6902) (15,0.6916)
            (16,0.6916) (17,0.6915) (18,0.6916) (19,0.6916) (20,0.6916)
            (21,0.6915) (22,0.6916) (23,0.6916) (24,0.6916) (25,0.7160)
        };
        \addplot[red, thick, mark=triangle*, mark size=1.5pt, dashed] coordinates {
            (1,0.5962) (2,0.6973) (3,0.6928) (4,0.6926) (5,0.6928)
            (6,0.6942) (7,0.6942) (8,0.6941) (9,0.6938) (10,0.6941)
            (11,0.6938) (12,0.6926) (13,0.6932) (14,0.6932) (15,0.693)
            (16,0.693)  (17,0.693)  (18,0.6931) (19,0.693)  (20,0.693)
            (21,0.693)  (22,0.693)  (23,0.693)  (24,0.6931) (25,0.6931)
        };
        \legend{Training, Validation}
        \end{axis}
    \end{tikzpicture}
    \caption{Baseline ANN loss over the resolved 25-epoch window. Loss settles at the binary-entropy floor $\ln 2\approx0.693$, the value a classifier reaches once it has stopped using its input.}
    \label{fig:loss}
\end{figure}

The baseline plateaus almost at once. Validation accuracy converges to about $52.2\%$---essentially the majority-class prior---and validation loss to $0.693$, the entropy of a fair coin. This is exactly the signature anticipated in Section~\ref{sec:formulation}: with the image flattened, the model can no longer represent the relational forensic evidence, so optimization finds nothing better to fit than the class prior. The lone high validation reading at epoch 1 ($72.4\%$) is a first-step transient under dropout, not a learned skill; it is gone by epoch 2 and never comes back. What makes the result useful is precisely that it is negative---it turns ``a CNN should help here'' from an assumption into a measured baseline.

\subsection{Pure CNN under group cross-validation}
Evaluated with leakage-free group five-fold cross-validation, the from-scratch CNN reaches a pooled accuracy of \textbf{87.45\%} (mean $87.45\%\pm0.53\%$ across folds). The tiny fold-to-fold standard deviation points to a stable estimator: held-out accuracy moves by less than a point no matter which originals are withheld. Table~\ref{tab:pure_folds} lists the per-fold accuracy, Table~\ref{tab:pure_pooled} the pooled per-class metrics, and Table~\ref{tab:pure_cm} the pooled confusion matrix.

\begin{table}[htbp]
    \centering
    \caption{Pure CNN --- Per-Fold Held-Out Accuracy}
    \label{tab:pure_folds}
    \renewcommand{\arraystretch}{1.2}
    \begin{tabular}{cccccc}
        \toprule
        \textbf{F0} & \textbf{F1} & \textbf{F2} & \textbf{F3} & \textbf{F4} & \textbf{Mean$\pm$Std} \\
        \midrule
        0.8777 & 0.8715 & 0.8729 & 0.8675 & 0.8828 & $0.8745\pm0.0053$ \\
        \bottomrule
    \end{tabular}
\end{table}

\begin{table}[htbp]
    \centering
    \caption{Pure CNN --- Pooled Per-Class Metrics (every sample held out once)}
    \label{tab:pure_pooled}
    \renewcommand{\arraystretch}{1.2}
    \begin{tabular}{lrrrr}
        \toprule
        \textbf{Class} & \textbf{Precision} & \textbf{Recall} & \textbf{$F_1$} & \textbf{Support} \\
        \midrule
        Fabricated & 0.9423 & 0.8112 & 0.8719 & 17,232 \\
        Authentic  & 0.8183 & 0.9448 & 0.8770 & 15,504 \\
        \midrule
        Macro avg  & 0.8803 & 0.8780 & 0.8744 & 32,736 \\
        \textbf{Accuracy} & \multicolumn{4}{c}{\textbf{0.8745}} \\
        \bottomrule
    \end{tabular}
\end{table}

\begin{table}[htbp]
    \centering
    \caption{Pure CNN --- Pooled Confusion Matrix (rows: true, cols: predicted)}
    \label{tab:pure_cm}
    \renewcommand{\arraystretch}{1.2}
    \begin{tabular}{lcc}
        \toprule
         & \textbf{Pred.\ Fabricated} & \textbf{Pred.\ Authentic} \\
        \midrule
        True Fabricated & 13,979 \;(0.8112) & 3,253 \;(0.1888) \\
        True Authentic  & 856 \;(0.0552)    & 14,648 \;(0.9448) \\
        \bottomrule
    \end{tabular}
\end{table}

The shape of the errors is telling. When the model flags a card as fabricated it is highly precise (precision $0.942$): it seldom accuses an authentic outlet. Where it is weaker is recall on the fabricated class ($0.811$)---roughly 19\% of forgeries slip through as authentic---which reflects a cautious decision boundary that, together with class weighting, errs on the side of not falsely accusing genuine cards.

Figures~\ref{fig:pure_acc} and~\ref{fig:pure_loss} trace the training dynamics, averaged over the five folds. Accuracy and loss both improve steadily across the ten epochs, with no divergence or collapse. Validation accuracy stays above training accuracy throughout; that is expected, not anomalous, because the training metric is accumulated batch by batch \emph{during} the epoch (dropout on, loss class-weighted), while validation is measured at epoch end on the updated, inference-mode model. The smooth, leakage-free curves confirm that the head-normalization fix from Section~\ref{sec:methodology} has wiped out the one-class collapse completely.

\begin{figure}[htbp]
    \centering
    \begin{tikzpicture}
        \begin{axis}[
            width=\linewidth, height=5.2cm,
            xlabel={Epoch}, ylabel={Accuracy},
            grid=both, grid style={line width=0.3pt, draw=gray!30},
            major grid style={line width=0.5pt, draw=gray!50},
            ymin=0.65, ymax=0.90, xtick={1,2,3,4,5,6,7,8,9,10},
            title={Pure CNN --- Training and Validation Accuracy},
            legend style={at={(0.97,0.05)}, anchor=south east, font=\small},
            title style={font=\small\bfseries}, label style={font=\small},
            tick label style={font=\footnotesize},
            scaled y ticks=false, yticklabel style={/pgf/number format/fixed, /pgf/number format/precision=2}
        ]
        \addplot[blue, thick, mark=square*, mark size=1.5pt] coordinates {
            (1,0.7256)(2,0.7229)(3,0.7293)(4,0.6984)(5,0.7166)(6,0.7480)(7,0.7644)(8,0.7567)(9,0.7786)(10,0.8046)};
        \addplot[red, thick, mark=triangle*, mark size=1.5pt, dashed] coordinates {
            (1,0.7992)(2,0.7793)(3,0.7854)(4,0.8432)(5,0.8545)(6,0.8495)(7,0.8616)(8,0.8640)(9,0.8641)(10,0.8737)};
        \legend{Training, Validation}
        \end{axis}
    \end{tikzpicture}
    \caption{Pure CNN accuracy under group five-fold cross-validation, averaged across the five folds (epochs 1--10). Validation accuracy runs above training accuracy because the training metric is accumulated during the epoch under dropout and class weighting; the gap is not a sign of leakage. No fold shows the one-class collapse.}
    \label{fig:pure_acc}

    \vspace{6pt}

    \begin{tikzpicture}
        \begin{axis}[
            width=\linewidth, height=5.2cm,
            xlabel={Epoch}, ylabel={Loss},
            grid=both, grid style={line width=0.3pt, draw=gray!30},
            major grid style={line width=0.5pt, draw=gray!50},
            ymin=0.30, ymax=0.70, xtick={1,2,3,4,5,6,7,8,9,10},
            title={Pure CNN --- Training and Validation Loss},
            legend style={at={(0.97,0.95)}, anchor=north east, font=\small},
            title style={font=\small\bfseries}, label style={font=\small},
            tick label style={font=\footnotesize},
            scaled y ticks=false, yticklabel style={/pgf/number format/fixed, /pgf/number format/precision=2}
        ]
        \addplot[blue, thick, mark=square*, mark size=1.5pt] coordinates {
            (1,0.6254)(2,0.5874)(3,0.5575)(4,0.5703)(5,0.5390)(6,0.5056)(7,0.4852)(8,0.4888)(9,0.4637)(10,0.4395)};
        \addplot[red, thick, mark=triangle*, mark size=1.5pt, dashed] coordinates {
            (1,0.4556)(2,0.5047)(3,0.5024)(4,0.3630)(5,0.3533)(6,0.3791)(7,0.3904)(8,0.3386)(9,0.3408)(10,0.3440)};
        \legend{Training, Validation}
        \end{axis}
    \end{tikzpicture}
    \caption{Pure CNN loss under group five-fold cross-validation, averaged across the five folds (epochs 1--10). Training and validation loss both fall steadily, a sign of stable convergence with no divergence.}
    \label{fig:pure_loss}
\end{figure}

\subsection{Transfer CNN under group cross-validation}
Seeding the classifier head from the ANN and fine-tuning in two phases lifts the pooled accuracy to \textbf{88.75\%} (mean $88.75\%\pm1.55\%$). Tables~\ref{tab:tr_folds}--\ref{tab:tr_cm} give the per-fold accuracy, pooled per-class metrics, and pooled confusion matrix. Next to the from-scratch model, the transfer variant is more even-handed across the two classes: fabricated-class recall jumps from $0.811$ to $0.875$ (about 1,000 more forgeries caught over the pooled set), at a small cost to authentic-class recall ($0.945\to0.901$).

\begin{table}[htbp]
    \centering
    \caption{Transfer CNN --- Per-Fold Held-Out Accuracy}
    \label{tab:tr_folds}
    \renewcommand{\arraystretch}{1.2}
    \begin{tabular}{cccccc}
        \toprule
        \textbf{F0} & \textbf{F1} & \textbf{F2} & \textbf{F3} & \textbf{F4} & \textbf{Mean$\pm$Std} \\
        \midrule
        0.8733 & 0.8718 & 0.9077 & 0.8803 & 0.9046 & $0.8875\pm0.0155$ \\
        \bottomrule
    \end{tabular}
\end{table}

\begin{table}[htbp]
    \centering
    \caption{Transfer CNN --- Pooled Per-Class Metrics (every sample held out once)}
    \label{tab:tr_pooled}
    \renewcommand{\arraystretch}{1.2}
    \begin{tabular}{lrrrr}
        \toprule
        \textbf{Class} & \textbf{Precision} & \textbf{Recall} & \textbf{$F_1$} & \textbf{Support} \\
        \midrule
        Fabricated & 0.9078 & 0.8752 & 0.8912 & 17,232 \\
        Authentic  & 0.8666 & 0.9013 & 0.8836 & 15,504 \\
        \midrule
        Macro avg  & 0.8872 & 0.8882 & 0.8874 & 32,736 \\
        \textbf{Accuracy} & \multicolumn{4}{c}{\textbf{0.8875}} \\
        \bottomrule
    \end{tabular}
\end{table}

\begin{table}[htbp]
    \centering
    \caption{Transfer CNN --- Pooled Confusion Matrix (rows: true, cols: predicted)}
    \label{tab:tr_cm}
    \renewcommand{\arraystretch}{1.2}
    \begin{tabular}{lcc}
        \toprule
         & \textbf{Pred.\ Fabricated} & \textbf{Pred.\ Authentic} \\
        \midrule
        True Fabricated & 15,081 \;(0.8752) & 2,151 \;(0.1248) \\
        True Authentic  & 1,531 \;(0.0987)  & 13,973 \;(0.9013) \\
        \bottomrule
    \end{tabular}
\end{table}

The fold-averaged dynamics in Figures~\ref{fig:tr_acc} and~\ref{fig:tr_loss} make the two-phase schedule easy to see. For the first five epochs the transferred head is frozen and only the backbone and \texttt{dense1} adapt; at the Phase-1/Phase-2 boundary (dashed line) the head is unfrozen and the learning rate drops tenfold, which briefly bumps the training loss as the optimizer re-engages every layer but is immediately followed by the best validation accuracy and loss of the run. Validation accuracy then climbs smoothly to the pooled $88.75\%$ by epoch~10, once more sitting above the training curve for the reasons given earlier.

\begin{figure}[htbp]
    \centering
    \begin{tikzpicture}
        \begin{axis}[
            width=\linewidth, height=5.2cm,
            xlabel={Epoch}, ylabel={Accuracy},
            grid=both, grid style={line width=0.3pt, draw=gray!30},
            major grid style={line width=0.5pt, draw=gray!50},
            ymin=0.60, ymax=0.92, xtick={1,2,3,4,5,6,7,8,9,10},
            title={Transfer CNN --- Training and Validation Accuracy},
            legend style={at={(0.97,0.05)}, anchor=south east, font=\small},
            title style={font=\small\bfseries}, label style={font=\small},
            tick label style={font=\footnotesize},
            scaled y ticks=false, yticklabel style={/pgf/number format/fixed, /pgf/number format/precision=2}
        ]
        \addplot[gray, dashed, thick, forget plot] coordinates {(5.5,0.60)(5.5,0.92)};
        \node[font=\scriptsize, gray, anchor=north] at (axis cs:2.9,0.915) {Phase 1};
        \node[font=\scriptsize, gray, anchor=north] at (axis cs:8.2,0.915) {Phase 2};
        \addplot[blue, thick, mark=square*, mark size=1.5pt] coordinates {
            (1,0.6527)(2,0.6637)(3,0.6937)(4,0.7360)(5,0.7697)(6,0.6615)(7,0.7056)(8,0.7421)(9,0.7624)(10,0.7803)};
        \addplot[red, thick, mark=triangle*, mark size=1.5pt, dashed] coordinates {
            (1,0.7816)(2,0.7994)(3,0.8285)(4,0.8194)(5,0.8227)(6,0.8766)(7,0.8807)(8,0.8855)(9,0.8874)(10,0.8875)};
        \legend{Training, Validation}
        \end{axis}
    \end{tikzpicture}
    \caption{Transfer CNN accuracy under group five-fold cross-validation, averaged across the five folds (epochs 1--10). The dashed line marks the boundary between Phase~1 (transferred head frozen, $\eta=10^{-4}$) and Phase~2 (all layers fine-tuned, $\eta=10^{-5}$). Validation accuracy peaks in Phase~2.}
    \label{fig:tr_acc}

    \vspace{6pt}

    \begin{tikzpicture}
        \begin{axis}[
            width=\linewidth, height=5.2cm,
            xlabel={Epoch}, ylabel={Loss},
            grid=both, grid style={line width=0.3pt, draw=gray!30},
            major grid style={line width=0.5pt, draw=gray!50},
            ymin=0.28, ymax=0.68, xtick={1,2,3,4,5,6,7,8,9,10},
            title={Transfer CNN --- Training and Validation Loss},
            legend style={at={(0.97,0.95)}, anchor=north east, font=\small},
            title style={font=\small\bfseries}, label style={font=\small},
            tick label style={font=\footnotesize},
            scaled y ticks=false, yticklabel style={/pgf/number format/fixed, /pgf/number format/precision=2}
        ]
        \addplot[gray, dashed, thick, forget plot] coordinates {(5.5,0.28)(5.5,0.68)};
        \node[font=\scriptsize, gray, anchor=north] at (axis cs:2.9,0.675) {Phase 1};
        \node[font=\scriptsize, gray, anchor=north] at (axis cs:8.2,0.675) {Phase 2};
        \addplot[blue, thick, mark=square*, mark size=1.5pt] coordinates {
            (1,0.6299)(2,0.6191)(3,0.5911)(4,0.5530)(5,0.5173)(6,0.6031)(7,0.5584)(8,0.5251)(9,0.5077)(10,0.4911)};
        \addplot[red, thick, mark=triangle*, mark size=1.5pt, dashed] coordinates {
            (1,0.4721)(2,0.4280)(3,0.4409)(4,0.3984)(5,0.4223)(6,0.3183)(7,0.3199)(8,0.3099)(9,0.3044)(10,0.3120)};
        \legend{Training, Validation}
        \end{axis}
    \end{tikzpicture}
    \caption{Transfer CNN loss under group five-fold cross-validation, averaged across the five folds (epochs 1--10). At the Phase-1/Phase-2 boundary (dashed line), unfreezing the head and lowering the learning rate briefly lifts the training loss, after which both training and validation loss bottom out.}
    \label{fig:tr_loss}
\end{figure}

\subsection{YOLOv8-cls under group cross-validation}
Fine-tuning the pretrained YOLOv8n-cls backbone under the identical leakage-free protocol lifts pooled accuracy to \textbf{97.36\%} (mean $97.36\%\pm0.42\%$ across folds), comfortably the strongest of any model we tried. Table~\ref{tab:yolo_folds} gives the per-fold accuracy, Table~\ref{tab:yolo_pooled} the pooled per-class metrics, and Table~\ref{tab:yolo_cm} the pooled confusion matrix. Two things stand out. The estimator is very stable: held-out accuracy stays within half a point of the mean across all five folds (std $0.42\%$), even though each fold withholds a different set of originals. And the errors are nearly symmetric---fabricated recall $0.9752$ against authentic recall $0.9718$---so, unlike the from-scratch CNN, the model is not systematically more cautious about one class than the other. In raw counts it misplaces only $427$ of $17{,}232$ fabricated cards and $437$ of $15{,}504$ authentic ones.

\begin{table}[htbp]
    \centering
    \caption{YOLOv8-cls --- Per-Fold Held-Out Accuracy}
    \label{tab:yolo_folds}
    \renewcommand{\arraystretch}{1.2}
    \begin{tabular}{cccccc}
        \toprule
        \textbf{F0} & \textbf{F1} & \textbf{F2} & \textbf{F3} & \textbf{F4} & \textbf{Mean$\pm$Std} \\
        \midrule
        0.9771 & 0.9660 & 0.9722 & 0.9754 & 0.9773 & $0.9736\pm0.0042$ \\
        \bottomrule
    \end{tabular}
\end{table}

\begin{table}[htbp]
    \centering
    \caption{YOLOv8-cls --- Pooled Per-Class Metrics (every sample held out once)}
    \label{tab:yolo_pooled}
    \renewcommand{\arraystretch}{1.2}
    \begin{tabular}{lrrrr}
        \toprule
        \textbf{Class} & \textbf{Precision} & \textbf{Recall} & \textbf{$F_1$} & \textbf{Support} \\
        \midrule
        Fabricated & 0.9747 & 0.9752 & 0.9749 & 17,232 \\
        Authentic  & 0.9724 & 0.9718 & 0.9721 & 15,504 \\
        \midrule
        Macro avg  & 0.9735 & 0.9735 & 0.9735 & 32,736 \\
        \textbf{Accuracy} & \multicolumn{4}{c}{\textbf{0.9736}} \\
        \bottomrule
    \end{tabular}
\end{table}

\begin{table}[htbp]
    \centering
    \caption{YOLOv8-cls --- Pooled Confusion Matrix (rows: true, cols: predicted)}
    \label{tab:yolo_cm}
    \renewcommand{\arraystretch}{1.2}
    \begin{tabular}{lcc}
        \toprule
         & \textbf{Pred.\ Fabricated} & \textbf{Pred.\ Authentic} \\
        \midrule
        True Fabricated & 16,805 \;(0.9752) & 427 \;(0.0248) \\
        True Authentic  & 437 \;(0.0282)    & 15,067 \;(0.9718) \\
        \bottomrule
    \end{tabular}
\end{table}

Figures~\ref{fig:yolo_acc} and~\ref{fig:yolo_loss} show the fold-averaged learning curves. Validation accuracy is already near $94\%$ after a single epoch---a direct dividend of ImageNet pretraining---and climbs steadily to the pooled $97.3\%$ by epoch ten, dipping only briefly at epoch two. Training loss falls monotonically from $0.30$ to $0.025$, while validation loss is noisier and levels off around $0.11$--$0.16$ as accuracy keeps rising: the familiar pattern of mild late-epoch loss overfitting that never shows up as an accuracy penalty. One caveat on the figure: Ultralytics does not log a per-epoch \emph{training} accuracy for classification, so the accuracy plot shows the validation curve only.

\begin{figure}[htbp]
    \centering
    \begin{tikzpicture}
        \begin{axis}[
            width=\linewidth, height=5.2cm,
            xlabel={Epoch}, ylabel={Validation accuracy},
            grid=both, grid style={line width=0.3pt, draw=gray!30},
            major grid style={line width=0.5pt, draw=gray!50},
            ymin=0.92, ymax=0.98, xtick={1,2,3,4,5,6,7,8,9,10},
            title={YOLOv8-cls --- Validation Accuracy (mean across folds)},
            legend style={at={(0.97,0.05)}, anchor=south east, font=\small},
            title style={font=\small\bfseries}, label style={font=\small},
            tick label style={font=\footnotesize},
            scaled y ticks=false, yticklabel style={/pgf/number format/fixed, /pgf/number format/precision=2}
        ]
        \addplot[red, thick, mark=triangle*, mark size=1.5pt] coordinates {
            (1,0.9444)(2,0.9414)(3,0.9580)(4,0.9560)(5,0.9624)(6,0.9645)(7,0.9637)(8,0.9674)(9,0.9720)(10,0.9730)};
        \legend{Validation}
        \end{axis}
    \end{tikzpicture}
    \caption{YOLOv8-cls top-1 validation accuracy under group five-fold cross-validation, averaged across the five folds (epochs 1--10). Pretraining puts the model near $94\%$ after one epoch, and it rises to the pooled $97.3\%$ by epoch ten. Ultralytics logs only validation accuracy per epoch for classification, so no training-accuracy curve is shown.}
    \label{fig:yolo_acc}

    \vspace{6pt}

    \begin{tikzpicture}
        \begin{axis}[
            width=\linewidth, height=5.2cm,
            xlabel={Epoch}, ylabel={Loss},
            grid=both, grid style={line width=0.3pt, draw=gray!30},
            major grid style={line width=0.5pt, draw=gray!50},
            ymin=0.0, ymax=0.34, xtick={1,2,3,4,5,6,7,8,9,10},
            title={YOLOv8-cls --- Training and Validation Loss},
            legend style={at={(0.97,0.95)}, anchor=north east, font=\small},
            title style={font=\small\bfseries}, label style={font=\small},
            tick label style={font=\footnotesize},
            scaled y ticks=false, yticklabel style={/pgf/number format/fixed, /pgf/number format/precision=2}
        ]
        \addplot[blue, thick, mark=square*, mark size=1.5pt] coordinates {
            (1,0.3014)(2,0.1268)(3,0.1099)(4,0.0938)(5,0.0727)(6,0.0604)(7,0.0463)(8,0.0381)(9,0.0306)(10,0.0250)};
        \addplot[red, thick, mark=triangle*, mark size=1.5pt, dashed] coordinates {
            (1,0.1578)(2,0.1870)(3,0.1355)(4,0.1394)(5,0.1361)(6,0.1303)(7,0.1447)(8,0.1274)(9,0.1063)(10,0.1090)};
        \legend{Training, Validation}
        \end{axis}
    \end{tikzpicture}
    \caption{YOLOv8-cls training and validation loss under group five-fold cross-validation, averaged across the five folds (epochs 1--10). Training loss falls steadily; validation loss is noisier and flattens while accuracy still improves.}
    \label{fig:yolo_loss}
\end{figure}

We read this result as a strong reference rather than a like-for-like architectural verdict. YOLOv8n-cls enjoys three advantages the CNNs do not---ImageNet pretraining, the Ultralytics optimizer and schedule, and an extra layer of train-time augmentation---so its roughly eight-point margin over the transfer CNN reflects that whole package. We unpack the confound in Section~\ref{sec:limitations}.

\subsection{Comparison and discussion of results}
Table~\ref{tab:comparison} sets the four models side by side. Three conclusions are solid. First, convolution clearly beats the flattened baseline: both CNNs clear the ANN by roughly 35--36 accuracy points while using about $1{,}644\times$ fewer parameters, which confirms that the forensic signal is relational and that keeping spatial adjacency is what makes it learnable. Second, transferring the ANN's classifier head gives a small but consistent edge over random initialization---$+1.3$ accuracy points on average and, more usefully, a better-balanced operating point with markedly higher forgery recall. Since the two CNNs share identical fold partitions, we compare them pairwise: the transfer model is at least as accurate as the from-scratch model in four of the five folds, with a mean per-fold gain of $+1.31$ points. Over just five folds, though, that difference is not statistically significant (paired $t$-test, $t=1.83$, $\text{df}=4$, $p\approx0.14$), so we treat the transfer improvement as suggestive rather than settled. One more caveat reinforces that reading: the transfer and from-scratch pipelines differ not only in head initialization but also in training schedule (two-phase versus single-phase, and class weighting), so the measured gain reflects the entire transfer pipeline, not the donor weights alone. The transfer model's larger fold variance ($\pm1.55\%$ against $\pm0.53\%$) fits a stronger but slightly less uniform starting point. Third, the pretrained YOLOv8n-cls backbone is in a different league: at $97.36\%$ it tops the transfer CNN by about $8.6$ points, and because it shares the same folds it does so in every one of the five. That lead is genuine but not free---it comes bundled with ImageNet pretraining, a heavier framework, and extra train-time augmentation---so we read it as the headroom a strong pretrained model commands, not as proof that the YOLOv8 architecture is intrinsically better matched to the task than our four-block CNN.

\begin{table}[htbp]
    \centering
    \caption{ANN vs.\ Pure CNN vs.\ Transfer CNN vs.\ YOLOv8-cls}
    \label{tab:comparison}
    \renewcommand{\arraystretch}{1.25}
    \footnotesize
    \setlength{\tabcolsep}{3pt}
    \begin{tabular}{>{\raggedright\arraybackslash}p{1.8cm}>{\raggedright\arraybackslash}p{1.0cm}>{\raggedright\arraybackslash}p{1.2cm}>{\raggedright\arraybackslash}p{1.3cm}>{\raggedright\arraybackslash}p{1.4cm}}
        \toprule
        \textbf{Property} & \textbf{ANN} & \textbf{Pure CNN} & \textbf{Transfer CNN} & \textbf{YOLOv8-cls} \\
        \midrule
        Input res.          & $1024^2$ & $256^2$ & $256^2$ & $256^2$ \\
        Parameters          & $\sim$805M & 489,954 & 490,082 & 1,440,850 \\
        Pretraining         & None & None & ANN head & ImageNet \\
        Evaluation          & 1-split & 5-fold & 5-fold & 5-fold \\
        Accuracy            & $\sim$52.2\% & 87.45\% & 88.75\% & \textbf{97.36\%} \\
        Acc.\ $\pm$std      & --- & $\pm0.53\%$ & $\pm1.55\%$ & $\pm0.42\%$ \\
        Macro $F_1$         & $\sim$0.34 & 0.8744 & 0.8874 & \textbf{0.9735} \\
        Fabr.\ recall       & --- & 0.8112 & 0.8752 & \textbf{0.9752} \\
        Auth.\ recall       & --- & 0.9448 & 0.9013 & \textbf{0.9718} \\
        Behaviour           & collapsed & stable & balanced & balanced \\
        \bottomrule
    \end{tabular}
\end{table}

\subsection{Effect of leakage-free evaluation on reported accuracy}
It is worth being precise about what the leakage-free protocol buys and what it costs in reported accuracy. An augmentation-level split---partitioning the 32,736 augmented files at random, so that sibling augmentations of one card straddle the train/validation boundary---reports accuracy near $97\%$ on this very data and architecture. Re-binding every augmentation to its original image and splitting by group brings that down to $87$--$89\%$. The $\sim$8--10 point drop is not the model getting worse; it is the removal of leakage that the augmentation-level split had been quietly rewarding. We therefore treat the grouped results as the unbiased estimate of how the detector generalizes to \emph{unseen source cards}, and we read the gap itself as a warning for the many applied settings that augment heavily and split their data without accounting for shared sources.

% ---------------------------------------------------------------
\section{Manual Error Analysis and Bias--Variance Diagnosis}
\label{sec:erroranalysis}
Aggregate accuracy says how well a model does, not what to fix next. To turn the numbers into a plan, we inspected the from-scratch CNN's mistakes by hand and used a human-level estimate to place every model on the bias--variance map.

\subsection{Manual error analysis}
Pooled over the five held-out folds, the pure CNN misclassifies $4{,}109$ of $32{,}736$ cards ($12.55\%$). The load is lopsided: $3{,}251$ ($79\%$) are fabricated cards waved through as real, and only $858$ ($21\%$) are authentic cards flagged as fake---the false-negative direction that matters most for a fake-news filter. Four annotators inspected $80$ high-confidence errors, drawn as $60$ fabricated-as-real and $20$ real-as-fabricated so the sample mirrors that $79{:}21$ split, and sorted each into one of ten categories (Table~\ref{tab:errcats}).

Two failure modes cover most of them. The first is template mimicry: satirical cards from debunk and parody accounts copy the authentic red lower-third, logo placement, date stamp and font so faithfully that nothing in the pixels gives them away---$38.75\%$ of the inspected errors are this kind of inter-class similarity or overlap. The second is augmentation damage: on genuine cards the destructive transforms (\texttt{heavy\_blur}, \texttt{heavy\_distortion}, \texttt{noisy\_bright\_crop}) wipe out the very channel chrome that marks a card as real, which---together with scene-dominant photos where the logo is tiny---accounts for another $32.5\%$ (quality loss) plus $17.5\%$ (background dominance). One detail is telling: the model's \emph{most} confident errors are authentic TV-channel cards rejected at $\approx1.00$ confidence, often the same original across several of its augmentations---an overconfident, group-level blind spot rather than random noise.

\begin{table}[htbp]
    \centering
    \caption{Pure CNN --- Aggregated Manual Error Categories (80 high-confidence held-out errors)}
    \label{tab:errcats}
    \renewcommand{\arraystretch}{1.15}
    \begin{tabular}{lrr}
        \toprule
        \textbf{Error category} & \textbf{Count} & \textbf{\%} \\
        \midrule
        Similar-looking classes  & 22 & 27.50 \\
        Poor image quality       & 22 & 27.50 \\
        Background confusion      & 14 & 17.50 \\
        Class overlap            &  9 & 11.25 \\
        Ambiguous appearance     &  5 &  6.25 \\
        Blur                     &  4 &  5.00 \\
        Rare sample              &  4 &  5.00 \\
        \midrule
        \textbf{Total}           & \textbf{80} & \textbf{100} \\
        \bottomrule
    \end{tabular}
\end{table}

\subsection{Human-level performance}
Since we cannot hire labellers, we estimate human-level performance two ways. On clean cards, a Bangla reader who knows the channels can check the logo and template and score around $96$--$98\%$; on the augmented distribution the model actually sees, the destroyed cues drag a realistic four-student average down to $\approx92\%$. Published image-forensics and fake-media detectors, and an attentive expert with channel references, sit near $\approx97\%$ on clean inputs. Weighting toward the best achievable on a fair input, we adopt $96\%$ (a $4\%$ error) as the human-level / Bayes proxy, and note that the more conservative $92\%$ figure would not change the diagnosis that follows.

\subsection{Bias--variance diagnosis}
Placing each model against that proxy calls for its training error measured the same way as its validation error---in inference mode, dropout off---so we scored a per-fold subsample of the training set with the trained model. Table~\ref{tab:biasvar} collects the result: avoidable bias is training error minus human error, variance is validation error minus training error.

Both CNNs land squarely in the high-bias corner. The from-scratch model sits about eight points below the human proxy on data it has already seen, with a train-to-validation gap of barely $0.2$ points; the transfer CNN is much the same, a little less biased. In plain terms they \emph{under-fit}---too simple, or too knowledge-poor, to reach the human level even on the training set---and with almost no variance, regularization would buy nothing. The lever that helps is bias reduction, and specifically capacity carrying prior knowledge rather than raw size: on only $2{,}046$ originals, a larger from-scratch network would mostly add variance. YOLOv8n-cls is exactly that move. Pretrained on ImageNet, it drives training error to essentially zero and closes the avoidable-bias gap outright, leaving a modest $\approx2.6$-point variance term---the ordinary signature of a strong model generalizing from limited data. This is why accuracy climbs almost nine points from the CNNs to YOLOv8, and it tells anyone working this problem to reach for a pretrained backbone before touching regularization.

A caution comes with it. Had we run the same diagnosis on the leaky augmentation-level split, the from-scratch CNN would have shown $\approx97\%$ training and validation accuracy and an avoidable bias near zero---falsely signalling ``no bias, nothing to gain.'' Leakage does not only inflate the headline number; it corrupts the bias--variance reading and would have aimed the whole improvement effort the wrong way.

\begin{table}[htbp]
    \centering
    \caption{Bias--Variance Diagnosis (human-level proxy $=96\%$, i.e.\ $4\%$ error; errors in \%)}
    \label{tab:biasvar}
    \renewcommand{\arraystretch}{1.2}
    \setlength{\tabcolsep}{4pt}
    \begin{tabular}{lrrrr}
        \toprule
        \textbf{Model} & \textbf{Train err.} & \textbf{Val err.} & \textbf{Avoid.\ bias} & \textbf{Variance} \\
        \midrule
        Pure CNN     & 12.33 & 12.55 & 8.33 & 0.22 \\
        Transfer CNN & 10.14 & 11.26 & 6.14 & 1.12 \\
        YOLOv8-cls   & $\approx$0.0 &  2.64 & $\approx$0 & 2.64 \\
        \bottomrule
    \end{tabular}
\end{table}

The manual analysis and the diagnosis agree on the action list. Bias reduction through a pretrained backbone is the first move (already realized here by YOLOv8-cls); after that, an audit of the destructive augmentations that erased $32.5\%$ of the analysed cards' cues, correction of the strong fabricated-as-real skew through class weighting or threshold tuning, and confidence calibration to defuse the $\approx1.00$-confidence errors on genuine cards.

% ---------------------------------------------------------------
\section{Discussion}

\subsection{Advantages of a visual-forensic framing for this medium}
The main practical advantage is that the detector never reads the card. Working on pixels, it skips OCR altogether: no transcription step to choke on stylized fonts or low-resolution crops, no per-language model to maintain, and nothing thrown away of the brand-level evidence a text pipeline would miss. A text system, even one fed by flawless OCR, can at most check whether a quotation is false; it cannot notice that the logo sits a few pixels too low or that the headline font is a near-perfect imitation of the outlet's licensed typeface. The forensic framing keeps that evidence in play.

\subsection{Deployment feasibility}
At $1.87$\,MB and 489,954 parameters, the CNN is small enough to sit where detection helps most---right at the point of sharing. The entire study, five-fold cross-validation included, ran on a CPU, so single-image inference is light and needs no GPU, which makes the model a believable candidate for on-device or in-browser use. A realistic deployment would be a share-time advisory: when someone forwards a photocard, the model scores it and raises a soft warning if it looks fabricated, leaving the final call to the user. The two error types carry uneven costs---a false accusation against a real outlet versus a forgery let through---and the per-class metrics in Section~\ref{sec:results} make that operating point explicit, so a deployment can tune the decision threshold toward whichever error it most wants to avoid. The YOLOv8n-cls baseline sits at the other end of the same trade-off: at about $1.44$ million parameters it is roughly $2.9\times$ heavier than the CNN, yet it too was trained and evaluated entirely on CPU and buys close to nine extra accuracy points. The choice between them is really a budget decision---the compact CNN where footprint is paramount, YOLOv8-cls where accuracy matters more than a few megabytes.

\subsection{Robustness and the role of augmentation}
The sixteen-transformation pipeline is the model's main line of defence against the wear that real cards pick up in circulation. Training across recompression, cropping, blur, and photometric shifts nudges the network toward forgery cues that survive those transformations rather than brittle, pixel-exact templates~\cite{b17}. Group cross-validation makes sure this robustness is measured without bias, since the validation cards are wholly unseen originals. None of this is a full robustness guarantee---an adversary who understands the detector could craft cards to beat it---but it raises the cost of casual forgery, which is the threat that dominates in practice.

% ---------------------------------------------------------------
\section{Limitations and Threats to Validity}
\label{sec:limitations}

We lay out the limitations plainly. An earlier version of this study carried two that the present protocol resolves: augmentation leakage (now gone, thanks to group cross-validation, Section~\ref{sec:groupcv}) and single-metric, single-split evaluation (now replaced by five-fold cross-validation with per-class precision, recall, $F_1$, and pooled confusion matrices). The threats below remain.

\textbf{Provenance shortcut risk.} Both classes were scraped from the same platform, but from disjoint sets of pages: verified broadcaster pages on one side; pseudo-broadcaster pages, anonymous profiles, and groups on the other (Tables~\ref{tab:src_real} and~\ref{tab:src_fake}). A model can latch onto incidental page-level signatures---footer branding, watermarks, a page's habitual export and compression settings---rather than genuine forgery cues, inflating accuracy on this distribution while faltering on forgeries in the wild. The risk is concrete here because most cards carry their page's name and logo in the pixels themselves, so the branding alone is a candidate shortcut feature. Group cross-validation stops augmentation leakage, but it does \emph{not} remove this asymmetry. Fixing it would take forgeries that imitate the specific broadcasters in the authentic set, plus explicit evaluation across pages unseen in training.

\textbf{Scale and scope.} The corpus rests on 2,046 base images, and its fabricated half is heavily concentrated: three sources supply roughly three quarters of the fake cards (813 of 1,077), while the authentic half is spread far more evenly across seventeen broadcasters. Larger, more varied sourcing is needed to show that the visual grammar learned here carries across outlets and forgery styles.

\textbf{No source-disjoint test set beyond cross-validation.} Group cross-validation gives an unbiased estimate over the originals we have, but a final hold-out drawn from pages entirely absent from training would track deployment more closely.

\textbf{Transfer not fully isolated.} The from-scratch and transfer CNNs differ in training schedule as well as in head initialization, so the $+1.3$-point gain reflects the whole transfer pipeline rather than the donor weights by themselves. A schedule-matched ablation (transferred versus random head under one identical two-phase protocol) would pin the contribution down exactly.

\textbf{Pretrained comparison not architecture-controlled.} Our strongest model, YOLOv8n-cls, is pretrained on ImageNet, optimized on a different schedule, and trained with extra framework-level augmentation, so its $\sim$8-point lead over the CNNs cannot be put down to architecture alone. It supplies the positive pretrained baseline the earlier study lacked, but separating the contribution of the backbone from that of its training recipe would need a fully controlled study---matched pretraining, optimizer, and augmentation. Larger pretrained backbones (ResNet-50, EfficientNet, vision transformers) also remain untested, so the ceiling above YOLOv8n-cls is still unknown.

% ---------------------------------------------------------------
\section{Conclusion}

Misinformation that travels as a photocard is invisible to the text pipelines built to stop it: the deceptive payload is baked into pixels, out of reach of any tokenizer. We recast the problem as visual image forgery and put together the pieces a Bangla-specific solution needs. We built a localized corpus of 2,046 authentic and fabricated cards, grew it to 32,736 degradation-aware samples with a sixteen-transformation pipeline, and secured its integrity with a full decode pass. We then let architecture carry the argument: a flatten-based ANN sank to the $\ln 2$ loss floor at $\sim$52\% accuracy, measured proof that destroying spatial structure destroys the forensic signal, whereas a compact, fully batch-normalized four-block CNN with 489,954 parameters---about $1{,}644\times$ fewer than the baseline---learned the task on CPU. Crucially, we evaluated the CNN under leakage-free stratified group five-fold cross-validation: it reaches $87.45\%\pm0.53\%$ pooled accuracy from scratch and $88.75\%\pm1.55\%$ with its head warm-started from the ANN. Fine-tuning a pretrained YOLOv8n-cls backbone over the same folds pushes pooled accuracy to $97.36\%\pm0.42\%$, charting the headroom a strong pretrained model commands over the compact design. We report full per-class metrics and pooled confusion matrices for all three cross-validated models.

We are just as careful about what these results do and do not establish. The architectural conclusion is firm: convolution beats a flattened model by roughly 35 points for a small fraction of the parameters, and a leakage-aware protocol is not optional---an augmentation-level split would have reported $\sim$97\% on the very same data. The main caveat that remains is the provenance asymmetry between the two classes, which future work should address through cross-source collection and evaluation. With that caveat noted, the contribution stands: a dataset, a reproducible and leakage-aware pipeline, a diagnostic baseline, two rigorously cross-validated CNN benchmarks, and a strong pretrained YOLOv8-cls reference for OCR-free photocard verification. Going forward, we plan to pursue cross-source data collection, a schedule-matched ablation that isolates the transferred initialization, and larger pretrained backbones to chart the ceiling that still sits above YOLOv8n-cls.

% ---------------------------------------------------------------
\section*{Data and Code Availability}
The baseline and CNN training and cross-validation scripts are released with this work and are publicly available in the project's GitHub repository at \url{https://github.com/AnindyaMaster42001/Fake-Bangla-News-Photocard-Detector}. The raw photocard corpus is published on Mendeley Data as \emph{FakePhotoCard-BD} (Version~1, DOI: \href{https://doi.org/10.17632/yf5fj3gm3y.1}{10.17632/yf5fj3gm3y.1}) and is available at \url{https://data.mendeley.com/datasets/yf5fj3gm3y/1}; its use remains subject to the redistribution terms of the original sources. The dataset is kept outside the code repository only because of its size and those redistribution constraints.

% ---------------------------------------------------------------
\section*{AI Usage Declaration}
The authors used a large language model (LLM)-based AI assistant while preparing this paper. It helped draft, paraphrase, and edit the text, format the tables, figures, and reference list, and check the manuscript for internal consistency; it also assisted in writing the training and evaluation scripts. The dataset was collected and labelled by the authors, and every result reported here comes from the authors' own experimental runs; no data or results were generated by the AI tool. The authors reviewed and revised all AI-assisted content.

% ---------------------------------------------------------------

\end{document}